\documentclass[preprint,12pt]{elsarticle}

\usepackage{amssymb}
\usepackage{graphicx}
\usepackage{latexsym}
\usepackage{cite}
\usepackage{booktabs}
\usepackage[misc]{ifsym}
\usepackage{xcolor}
\usepackage{hyperref}
\usepackage{natbib}
\usepackage{amssymb}
\usepackage{url}
\usepackage{hyperref}
\usepackage{multirow}
\usepackage{epsfig}
\usepackage{mathrsfs}
\usepackage{multirow}
\usepackage{algorithm}
\usepackage{algorithmic}
\usepackage{makecell}

\journal{arxiv.org}

\begin{document}

\begin{frontmatter}



\title{A Novel Garment Transfer Method Supervised by Distilled Knowledge of Virtual Try-on Model}


\author[label1]{Naiyu Fang}

\author[label1]{Lemiao Qiu\corref{cor1}}
\ead{qiulm@zju.edu.cn}

\author[label1]{Shuyou Zhang}

\author[label1]{Zili Wang}

\author[label1]{Kerui Hu}

\author[label1]{Jianrong Tan}

\affiliation[label1]{organization={State Key Laboratory of Fluid Power \& Mechatronic Systems, Zhejiang University},
            city={Hangzhou},
            postcode={310027},
            country={China}}

\cortext[cor1]{Corresponding author}
\begin{abstract}
Garment transfer can wear the garment of the model image onto the personal image. As garment transfer leverages wild and cheap garment input, it has attracted tremendous attention in the community and has a huge commercial potential. Since the ground truth of garment transfer is almost unavailable in reality, previous studies have treated garment transfer as either pose transfer or garment-pose disentanglement, and trained garment transfer in self-supervised learning, However, these implementation methods do not cover garment transfer intentions completely and face the robustness issue in the testing phase. Notably, virtual try-on technology has exhibited superior performance using self-supervised learning, we propose to supervise the garment transfer training via knowledge distillation from virtual try-on. Specifically, the overall pipeline is first to infer a garment transfer parsing, and to use it to guide downstream warping and inpainting tasks. The transfer parsing reasoning model learns the response and feature knowledge from the try-on parsing reasoning model and absorbs the hard knowledge from the ground truth. The progressive flow warping model learns the content knowledge from virtual try-on for a reasonable and precise garment warping. To enhance transfer realism, we propose an arm regrowth task to infer exposed skin. Experiments demonstrate that our method has state-of-the-art performance in transferring garments between persons compared with other virtual try-on and garment transfer methods.
\end{abstract}




\end{frontmatter}



\section{Introduction}
\label{sec1}

\begin{figure}[!ht]
\centering
\includegraphics[width=3.3in]{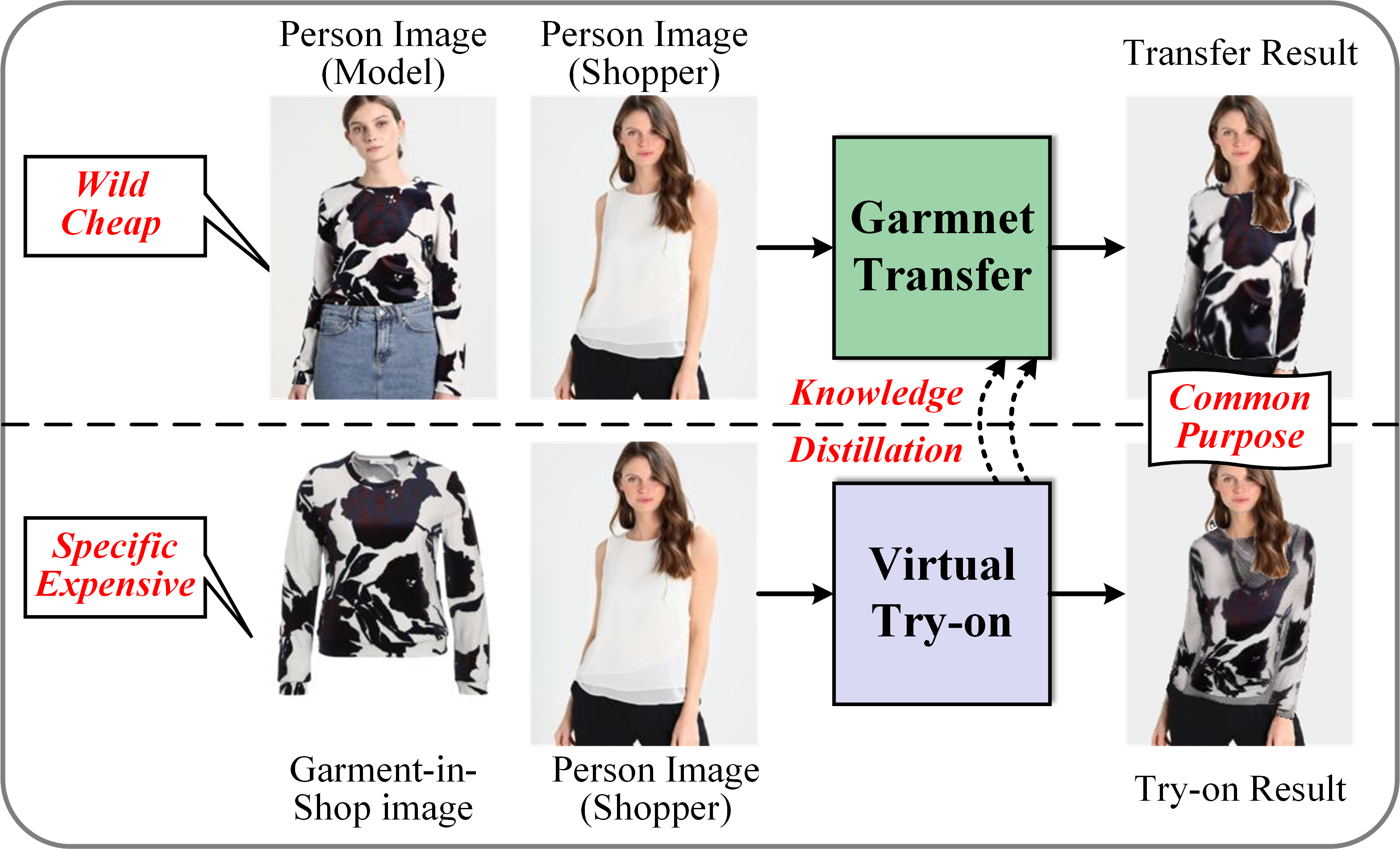}
\caption{Garment transfer vs virtual try-on.}
\label{fig1}
\end{figure}

Virtual try-on wears the garment-in-shop into a person, while garment transfer wears the garment from one person to another. As {\color{blue} Fig. \ref{fig1}} shows, the garment-in-shop of virtual try-on is captured in a controlled environment with professional equipment, while the personal image of garment transfer is wilder and cheaper. In a word, garment transfer also allows shoppers to perceive the garment-wearing effect but does not require specific and expensive garment conditions. However, the pair of garment-in-shop and the try-on result is easy to collect, and the pair of the person image and the transfer result is almost unavailable. Thus, garment transfer is a user-friendly but training-hard topic. As virtual try-on can be implemented by self-supervised learning and both technologies share the same purpose, we propose to supervise the training of garment transfer by distilling knowledge from virtual try-on. To the best of our knowledge, this is the first time to implement garment transfer via knowledge distillation. In this paper, we present a novel garment transfer pipeline and knowledge distillation process by exploiting previous virtual try-on studies.

To wear a garment-in-shop, the virtual try-on method warps the garment-in-shop towards the personal pose and preserves the other personal features. It involves many topics, such as image deformation, image inpainting, and image fusion. VITON {\color{blue}\citep{han2018viton}} proposed the concept of image-based virtual try-on for the first time and provided an inspiring implementation pipeline. It synthesizes a coarse try-on result, warps the garment-in-shop toward this result, and combines the synthesized and warped results to refine the garment texture. Alternatively, CP-VTON {\color{blue}\citep{wang2018toward}} offers a different pipeline where the warping and synthesis sequences are reversed. Since the paired garment-in-shop and the try-on result are available, they were both trained in a self-supervised manner by wearing a garment-in-shop onto a person dressed in the same garment. This self-supervision learning enables models to have perfect warping and inpainting capacities for general cases.

Intuitively, the garment-in-shop is nearly aligned and has a high pixel ratio since it is captured at a hanging status. Conversely, the garment in the person image is entangled with the personal pose, presenting challenges for the garment transfer model. When two persons wear different garments with different poses, the ground truth of garment transfer is one person wears the garment of another person while maintaining itself pose. However, it is almost impossible to collect such ground truth in reality. Therefore, the research key point of garment transfer is how to supervise its learning, and there are two dominant ways. One approach {\color{blue}\citep{raj2018swapnet}} is to disentangle the garment and pose in the personal image and re-entangle the pose and spatial-agnostic garment to generate the personal image. The other approach {\color{blue}\citep{yang2021ct}} is to directly alter the pose feature in the garment by warping. With a pair of person images where a person wears the same garment but has different poses, the model is trained to warp the garment from one pose to another.

With the easy-to-access dataset and easy-to-follow pipeline, virtual try-on studies are more flourishing than garment transfer studies. Existing garment transfer studies encounter several issues that hinder their ability to achieve transfer results comparable to virtual try-on studies. 1) The implementation method does not completely conform to its intention. It is imperative to disentangle pose from garment and results in high-frequency details in re-entangling the garment and pose. Training garment transfer as pose transfer within the same person faces shape and style preservation issues in transferring a garment to a different person; 2) The multi-task execution sequence is unreasonable. Garment warping is guided by a target pose instead of a transfer parsing, while the transfer parsing is reasoned after the garment warping task to guide final image synthesis. It is inevitable to yield misalignment and artifacts. 3) The specialized preservation mechanism of personal features receives inadequate study. It needs direct propagation and content inference to preserve the face and upper body skin. They just combine the warped garment and source personal features and result in artifacts in the new exposure part without a specialized inference mechanism.

To address these issues, we propose a novel garment transfer method where the training is supervised by virtual try-on via knowledge distillation. It just adopts the existing virtual try-on dataset and does not require paired person images. It is trained in a fully supervised manner so that the input conditions of the training and testing phase are identical. And knowledge distillation is employed during training transfer parsing reasoning and garment warping. We summarize the main contributions of our paper as follows:

(1) Teaching transfer parsing reasoning in multi-phase. The transfer parsing reasoning model learns the response and feature knowledge from the pre-trained try-on parsing reasoning model. We then transfer the garment back to its originator so that the transfer parsing reasoning model absorbs the hard knowledge from the ground truth to improve its robustness.

(2) Supervising garment warping with the shape and content information. We first map the garment to a similar position to the target shape. With the warping knowledge of garment-in-shop, a progressive flow is estimated to precisely warp the garment by learning the correspondence at both the shape and content level.

(3) Experiments demonstrate that our method has state-of-the-art performance compared to other virtual try-on and garment transfer methods in garment transfer. Our method performs well in transfer parsing reasoning, garment warping, new skin inference, and final result synthesis whatever the dressing and pose of person images are.

\section{Related Work}
\label{sec2}

\textbf{Virtual Try-on} VITON {\color{blue}\citep{han2018viton}} and CP-VTON {\color{blue}\citep{wang2018toward}} determined the fundamental strategy to implement virtual try-on, and the subsequent studies focused on optimizing garment warping and person reservation. Yang H et al. introduced second-order constraint into the TPS deformation in ACGPN {\color{blue}\citep{yang2020towards}}. Minar M R et al. proposed CP-VTON+ {\color{blue}\citep{minar2020cp}} extended on CP-VITON where the skin label is updated and the garment-in-shop mask is concatenated into the image synthesis. Neuberger A et al. {\color{blue}\citep{neuberger2020image}} proposed an online optimization step to optimize fine details reservation. Ge C et al. introduced cycle consistency to implement self-supervision learning in DCTON {\color{blue}\citep{ge2021disentangled}}. However, the coarse-grained TPS deformation used by these methods inevitably results in misalignment between the garment and person, requiring warping and rendering to be combined, which can introduce artifacts into the final result. To address this issue, some studies have used optical flow with higher DOF for finer warping. Ge Y et al. proposed PF-AFN {\color{blue}\citep{ge2021parser}} to estimate appearance flow from coarse to fine. Fang N et al. {\color{blue}\citep{fang2024pg}} proposed a warping-mapping-composition mechanism. He S et al. {\color{blue}\citep{he2022style}} proposed Flow Style VTON (FSV) to estimate a global warping tendency via style modulation based on PF-AFN. Additionally, some studies have used knowledge distillation to make their models independent of parsing and pose. Issenhuth T et al. proposed WUTON {\color{blue}\citep{issenhuth2020not}} to transfer warping knowledge and disambiguate the human parser error influence. PF-AFN {\color{blue}\citep{ge2021parser}} and FSV {\color{blue}\citep{he2022style}} linked the teacher model output with the student model input to implement the parser-free virtual try-on.

\textbf{Garment Transfer} It is regarded as garment-pose disentanglement or pose transfer for self-supervised learning. For the former, as the originator of garment transfer, SwapNet {\color{blue}\citep{raj2018swapnet}} learned the categories embedding by ROI pooling to disentangle the garment with the pose. Xie Z et al. proposed PASTA-GAN {\color{blue}\citep{xie2021towards}} to cut the garment into multiple patches for the disentanglement and exploited two StyleGAN2 to re-entangle garment patches with pose while considering style and texture. For the latter, Yang F et al. proposed two complementary warping in CT-Net {\color{blue}\citep{yang2021ct}} to implement garment pose changing. Liu T et al. proposed SPATT {\color{blue}\citep{liu2021spatial}} to establish a spatial correspondence in UV space to facilitate garment warping and unobserved region inference.

\textbf{Knowledge Distillation} It was proposed by Hinton G et al. in {\color{blue}\citep{hinton2015distilling}}, where the pre-trained teacher model supervises the student model learning by teaching various knowledge. This approach aims to compress the parameter quantity and training data, enabling the student model to be lightweight. The distilled knowledge can be categorized into three types: response knowledge, feature knowledge, and relation knowledge. Specifically, response knowledge {\color{blue}\citep{hinton2015distilling}} and feature knowledge {\color{blue}\citep{zagoruyko2017paying}} are extracted at the last and hint layer, containing prediction and processing information. And relation knowledge {\color{blue}\citep{ding2022dual}} describes the relationship between data or feature maps. The methods of knowledge distillation are categorized into offline distillation, online distillation, and self-distillation. Offline distillation employs a large model pre-trained on a large dataset as the teacher model and teaches a lightweight student model. In contrast, online distillation {\color{blue}\citep{liu2022self}} synchronously trains the teacher and student models to achieve better knowledge transfer when the large model is unavailable for specific tasks. Because of its universality and flexibility, knowledge distillation has been applied in several domains, including multi-view learning {\color{blue}\citep{tian2022multi}} and speaker extraction {\color{blue}\citep{huang2022compressing}}.

\section{Methodology}
\label{sec3}

\subsection{Outline}
\label{sec3.1}
We describe that the garment is transferred from person ${A}$ to person ${B}$. Assuming that the image of person ${A}$ and ${B}$ are ${{{\cal I}_A}}$ and ${{{\cal I}_B}}$, the corresponding garment-in-shop ${A}$ and ${B}$ are ${{\cal I}_A^c}$ and ${{\cal I}_B^c}$, virtual try-on intends to wear ${{\cal I}_B^c}$ onto ${{{\cal I}_A}}$, while garment transfer intends to transfer garment ${{\cal I}_B^{hc}}$ from ${{{\cal I}_B}}$ to ${{{\cal I}_A}}$. Since it is easy to collect the paired data ${\left( {{{\cal I}_A},{\cal I}_A^c} \right)}$, the virtual try-on model can be trained in a self-supervised manner. Conversely, the paired data ${\left( {{{\cal I}_A},{{\cal I}_B},{{\cal I}_{AB}}} \right)}$ is almost unavailable in the real world, it is a rock-hard issue to supervise the training of garment transfer. As they share a common objective, we endeavor to conduct the training of garment transfer by distilling knowledge from virtual try-on.

\begin{figure}[!ht]
\centering
\includegraphics[width=4.7in]{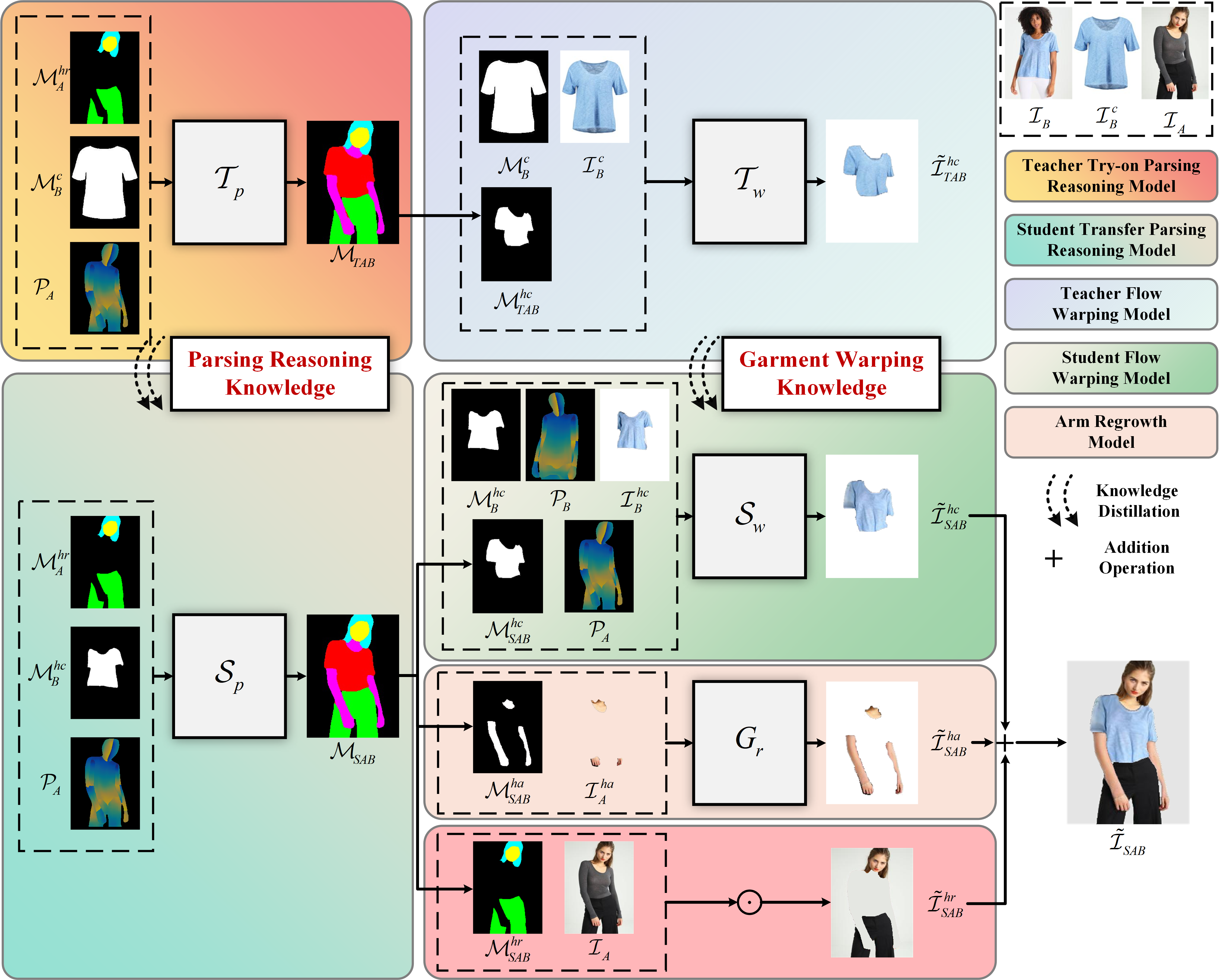}
\caption{The outline of the proposed method. We distill the parsing reasoning and garment warping knowledge in parsing reasoning and garment warping. Following training, ${{{\cal S}_p}}$ is capable of reasoning ${{{\cal M}_{SAB}}}$ to guide the downstream tasks, and ${{{\cal S}_w}}$ estimates a refined flow that warps ${{\cal I}_B^{hc}}$ towards ${{\cal M}_{SAB}^{hc}}$. We concatenate ${\tilde {\cal I}_{SAB}^{hc}}$, ${\tilde {\cal I}_{SAB}^{ha}}$, and ${\tilde {\cal I}_{SAB}^{hr}}$ to yield final transfer result ${{\tilde {\cal I}_{SAB}}}$.}
\label{fig2}
\end{figure}

As {\color{blue} Fig. \ref{fig2}} shows, we exploit the same pipeline for virtual try-on and garment transfer: parsing reasoning, garment warping, arm regrowth, and remaining content propagation. Virtual try-on teaches garment transfer via knowledge distillation in the parsing reasoning and garment warping tasks. For virtual try-on, the teacher transfer parsing reasoning model ${{{\cal T}_p}}$ and teacher flow warping model ${{{\cal T}_w}}$ are trained in a self-supervised manner. In the testing phase, ${{{\cal T}_p}}$ reasons the try-on parsing ${{{\cal M}_{TAB}}}$ where person ${A}$ wears the garment-in-shop ${B}$. Guided by the garment mask ${{\cal M}_{TAB}^{hc}}$ of ${{{\cal M}_{TAB}}}$, ${{{\cal T}_w}}$ warps ${{\cal I}_B^c}$ towards ${{\cal M}_{TAB}^{hc}}$ to yield ${\tilde {\cal I}_{TAB}^{hc}}$. For garment transfer, the student transfer parsing reasoning model ${{{\cal S}_p}}$ and the student flow warping model ${{{\cal S}_w}}$ are trained with supervision information distilled from ${{{\cal T}_p}}$ and ${{{\cal T}_w}}$. ${{{\cal S}_p}}$ reasons the transfer parsing ${{{\cal M}_{SAB}}}$ where person ${A}$ wears the garment of person ${B}$ ({\color{blue} Sec. \ref{sec3.2}}). With this target shape ${{\cal M}_{SAB}^{hc}}$, ${{{\cal S}_w}}$ warps garment ${{\cal I}_B^{hc}}$ to yield ${\tilde {\cal I}_{SAB}^{hc}}$ ({\color{blue} Sec. \ref{sec3.3}}). For the transfer realism, the arm regrowth model ${{G_r}}$ reasons more upper body skin ${\tilde {\cal I}_{SAB}^{ha}}$ when ${{\cal M}_{SAB}^{ha} > {\cal M}_A^{ha}}$ ({\color{blue} Sec. \ref{sec3.4}}). And for the remaining content, we just need inherit its pixel-level content from person ${A}$, via Boolean multiplying ${{\cal M}_{SAB}^{hr}}$ with ${{{\cal I}_A}}$. In the end, we concatenate ${\tilde {\cal I}_{SAB}^{ha}}$, ${\tilde {\cal I}_{SAB}^{hr}}$, and ${\tilde {\cal I}_{SAB}^{hc}}$ to obtain ${\tilde {\cal I}_{SAB}^{}}$.

To facilitate readability, we present our symbol naming conventions in the following enumeration:

$\bullet$ Sign. ${{\cal T}}$ and ${{\cal S}}$ are the teacher and student models; ${{\cal I}}$, ${{\cal M}}$, and ${{\cal P}}$ are the image, mask (parsing), and pose.

$\bullet$ Subscript. ${A}$ and ${B}$ represent the attribution to person ${A}$ and ${B}$; ${AB}$ represents the attribution to a new person ${A}$ who wears a garment identical to person ${B}$; ${S}$ and ${T}$ represent the result predicted by the student model and teacher model; ${p}$ and ${w}$ are the parsing reasoning model and flow warping model.

$\bullet$ Superscript. ${c}$ is the garment-in-shop; ${hc}$, ${ha}$, and ${hr}$ are the garment, upper body skin, and remaining category for a person; ${ \sim }$ represents a predicted result whose ground truth is the symbol without ${ \sim }$.

\subsection{The Transfer Parsing Reasoning Teaching}
\label{sec3.2}

\subsubsection{Parsing Reasoning Insight}
\label{sec3.2.1}

Garment transfer is a complex computer vision task that attempts to transfer garments from person ${B}$ to another person ${A}$, involving various shape-level and content-level tasks. To alleviate the computational burden of a single model to conduct these multi-level tasks, we prioritize the transfer parsing reasoning at the shape level, as it offers the critical shape guidance required for downstream tasks. This transfer parsing has the garment-agnostic features of person ${A}$ and the transferred garment features of person ${B}$.

The fundamental challenge of transfer parsing reasoning lies in how to effectively entangle the garment feature of person ${B}$ with the pose feature of person ${A}$, such that the garment conforms to the global pose feature of person ${A}$ while preserving its original style and category. This presents a significant challenge for both traditional fully supervised learning and self-supervised learning approaches. In the case of fully supervised learning, it is essential to acquire ground truth where person ${A}$ wears the garment of person ${B}$, but such data is rarely available in reality. Alternatively, in the case of self-supervised learning, the garment features undergo pose augmentation and the model is trained to entangle this augmented garment feature with its original pose feature. However, as data augmentationis insufficient to disentangle the original pose feature from the garment, the entanglement learning is prone to degenerate the plain mapping, leading to unreliable reasoning during the testing phase.

In a similar vein, albeit with greater accessibility, virtual try-on wears the garment-in-shop on a person, and we introduce a corresponding try-on parsing reasoning task. As garment-in-shop does not have any inherent pose features, through the exploitation of paired person parsing and garment-in-shop mask data, the entanglement of garment features and pose features can be effectively learned via self-supervised learning.

Assuming the parsing of person ${A}$ and ${B}$ are denoted as ${{{\cal M}_A}}$, ${{{\cal M}_B}}$, and their corresponding garment-in-shop masks are represented as ${{\cal M}_A^c}$, ${{\cal M}_B^c}$, we employ the try-on parsing reasoning model ${{{\cal T}_p}}$ and the transfer parsing reasoning model ${{{\cal S}_p}}$ as the teacher and student models, respectively. Specifically, we train the teacher model in a self-supervised manner to reason the try-on parsing ${{{\cal M}_{TAB}}}$. Thereafter, we employ knowledge distillation to train the student model to perform transfer parsing ${{{\cal M}_{SAB}}}$.

\subsubsection{Teacher Try-on Parsing Reasoning Model}
\label{sec3.2.2}

We exploit the pre-trained Grapy-ML {\color{blue}\citep{he2020grapy}} to extract person parsing ${{{\cal M}_A}}$, ${{{\cal M}_B}}$ from images ${{{\cal I}_A}}$, ${{{\cal I}_B}}$. To facilitate subsequent processing, the person parsing categories are grouped into background, hair, face, upper body skin, upper garment, lower garment, and thigh. Additionally, the pre-trained Densepose model {\color{blue}\citep{guler2018densepose}} is utilized to extract the dense pose features ${{{\cal P}_A}}$, ${{{\cal P}_B}}$ from images ${{{\cal I}_A}}$, ${{{\cal I}_B}}$. We adopt U2-Net as the backbone of ${{{\cal T}_p}}$ , which is proficient in capturing multiscale contextual information and is well-suited for the given reasoning task.

As described in {\color{blue} Sec. \ref{sec3.2.1}}, the objective of ${{{\cal T}_p}}$ is to entangle shape features of a new garment-in-shop with pose features of the person while also propagating garment-agnostic features of the person. It necessitates providing independent priors for each feature for this task. Accordingly, we leverage ${{\cal M}_A^c}$ and ${{{\cal P}_A}}$ to represent the garment shape and person pose, respectively. Since the upper body skin and background categories entangle with the garment category, including them in garment-agnostic features will hinder entanglement learning. Therefore, we exclude these categories from the parsing ${{{\cal M}_A}}$ and merge the remaining ones into ${{\cal M}_A^{hr}}$, which depicts the garment-agnostic shape in person parsing. During the self-supervised training, we concatenate ${{\cal M}_A^c}$,${{{\cal P}_A}}$,${{\cal M}_A^{hr}}$ and feed them into the model ${{{\cal T}_p}}$ to predict ${{\tilde {\cal M}_A}}$, which is supervised with the ground truth ${{{\cal M}_A}}$. During testing, we substitute ${{\cal M}_A^c}$ with ${{\cal M}_B^c}$ to predict ${{{\cal M}_{TAB}}}$, where person ${A}$ wears the garment-in-shop ${B}$. We denote this process as ${{{\cal M}_{TAB}} = {{\cal T}_p}\left( {{\cal M}_B^c,{{\cal P}_A},{\cal M}_A^{hr}} \right)}$ in the formula.

\subsubsection{Student Transfer Parsing Reasoning Model}
\label{sec3.2.3}
Once the teacher model has been trained, the teacher model ${{{\cal T}_p}}$ can impart knowledge for supervising the training of the student model ${{{\cal S}_p}}$. In this section, we will describe how to implement this teacher-teaches-student process.

Typically, when a teacher teaches a student how to solve a problem, they first introduce the basic knowledge and mechanism, followed by a solution procedure from implicit to explicit, and provide feedback to supervise the student's learning. If the student aims to be skilled in this type of problem, they need to do more exercises in self-study. Drawing inspiration from this, we delicately design the knowledge distillation between parsing reasoning models as three phases: feature knowledge teaching, response knowledge teaching, and self-study phase.

\begin{figure}[!ht]
\centering
\includegraphics[width=3.5in]{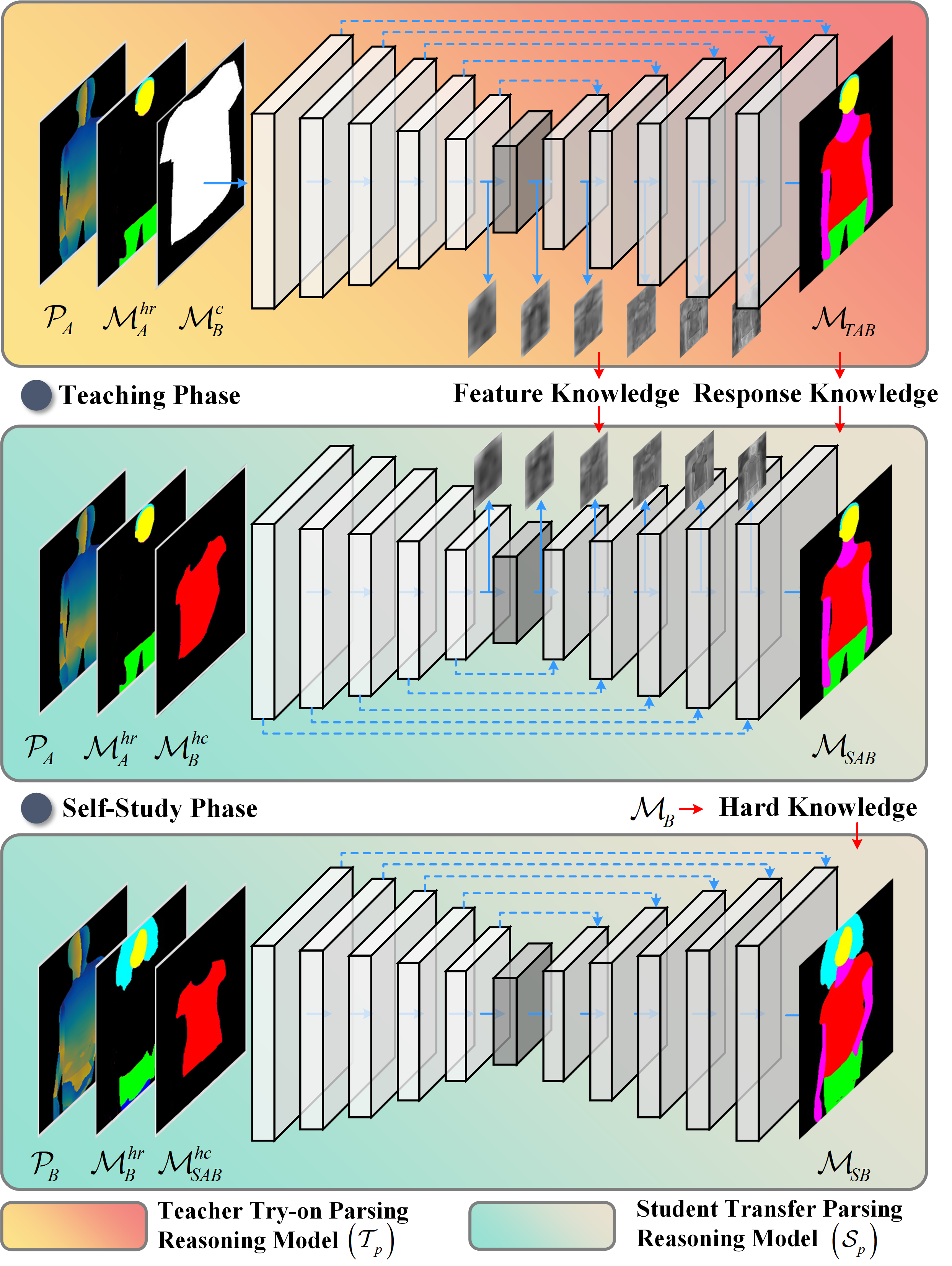}
\caption{Transfer parsing reasoning teaching. The pre-trained teacher model provides the feature and response knowledge to supervise the training of student model. The student model then advances its reasoning capability by learns the hard knowledge in the self-study phase.}
\label{fig3}
\end{figure}

We adopt the same backbone in ${{{\cal S}_p}}$ as in ${{{\cal T}_p}}$. During the feature and response knowledge teaching phase, ${{{\cal S}_p}}$ is conditioned on ${{\cal M}_B^{hc}}$, ${{{\cal P}_A}}$, and ${{\cal M}_A^{hr}}$. Due to the unaligned pose of person ${B}$ in ${{\cal M}_B^{hc}}$ compared to ${{\cal M}_B^c}$, transfer parsing reasoning poses a greater challenge compared to try-on parsing reasoning. As {\color{blue} Fig. \ref{fig3}} shows, the encoder in the backbone is responsible for feature extraction and position awareness. However, with the input difference between the teacher and student models, their learned knowledge in the encoder may not be identical. On the other hand, the decoder in the backbone is responsible for entangling semantic features and reconstructing the parsing structure, the teacher and student models share the same objective for it.. Therefore, we propose to distill the decoder's knowledge from ${{{\cal T}_p}}$ to supervise the training of ${{{\cal S}_p}}$.

\begin{figure}[!ht]
\centering
\includegraphics[width=3in]{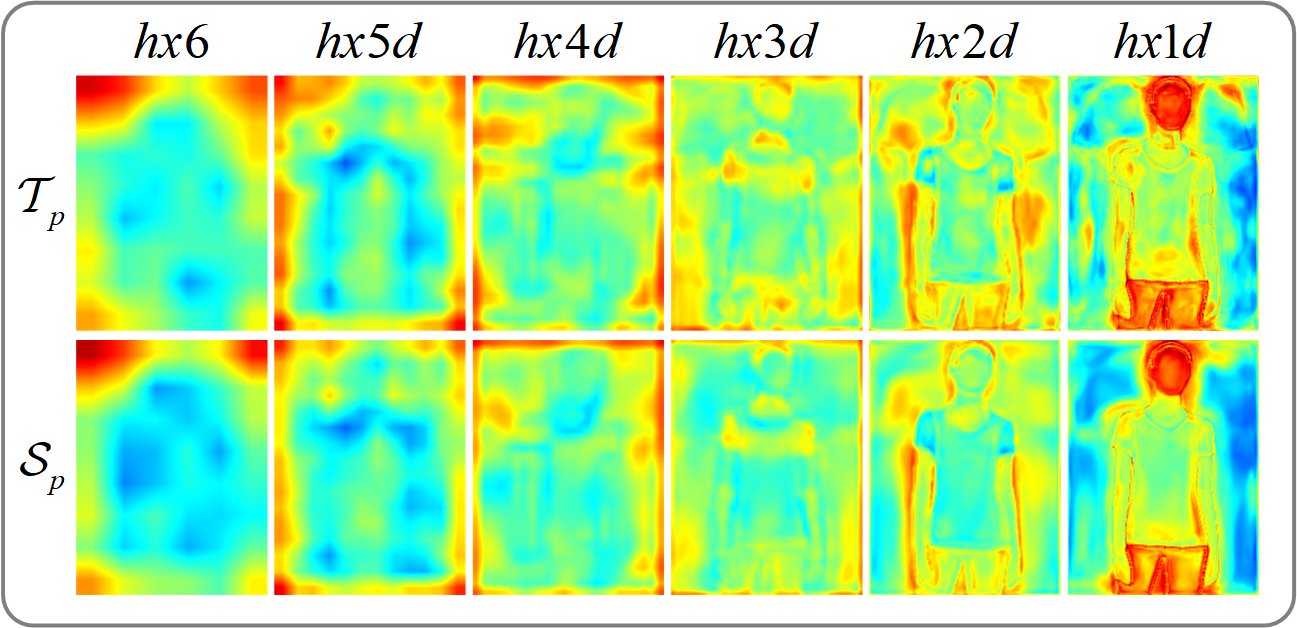}
\caption{The heat maps of feature knowledge.}
\label{fig4}
\end{figure}

Specifically, we focus on the layers 'hx6', 'hx5d', 'hx4d', 'hx3d', 'hx2d', and 'hx1d' \footnote{These signs are derived from the official U2-Net code (https://github.com/NathanUA/U-2-Net).} in the decoder, which output feature maps ranging from coarse to fine. As {\color{blue} Fig. \ref{fig4}} shows, it indicates that the global pose and overall shape are learned at the early layers, 'hx6' and 'hx5d', which are considered basic knowledge for reasoning. Subsequently, the model starts to re-establish human structure and distinguish categories at the intermediate layers, 'hx4d', 'hx3d', and 'hx2d', with particular attention to the boundary and overall shape of the garment and upper body skin, which represent the entanglement of the garment and pose features. Finally, at layer 'hx1d', our models refine the shape of garment-agnostic categories by fusing low-level shape information from skip connections. During the first stage, we only distilled knowledge from layers 'hx6', 'hx5d', which indirectly supervises ${{{\cal S}_p}}$ to extract hidden knowledge of garment style and target pose. During the second stage, we distilled knowledge from all layers, which supervises ${{{\cal S}_p}}$ at the entanglement of semantic features and reconstruction of parsing spatial relationships.

During the response knowledge teaching phase, we distill soft knowledge in the form of a heat map instead of one-hot coding. Since the soft target represented by ${{{\cal M}_{TAB}}}$ is not equivalent to the ground truth, ${{{\cal S}_p}}$ is able to acquire hard knowledge through self-study after the teaching phase, thereby improving its overall reasoning capacity and facilitating the correction of teaching errors. Specifically, we split the garment category mask ${{\cal M}_{SAB}^{hc}}$ from ${{{\cal M}_{SAB}}}$, re-transfer it onto person ${B}$, and supervise the reasoning result by the ground truth ${{{\cal M}_B}}$. It is described as ${{{\cal M}_{SB}} = {{\cal S}_p}\left( {{{\cal P}_B},{\cal M}_B^{hr},{\cal M}_{SAB}^{hc}} \right)}$ in the formula.

\subsubsection{Teaching Procedure}
\label{sec3.2.4}
{\color{blue} Algorithm \ref{algorithm1}} summarizes the teaching procedure of reasoning transfer parsing. To distinguish the teaching phase, we employ epoch intervals and utilize distinct combinations of objective functions ${{\ell _p}}$ across the various phases. Specifically, ${{\ell _{pf}}}$ represents the feature loss, defined as {\color{blue} (\ref{eq1})}, where ${\alpha _T^i}$ and ${\alpha _S^i}$ correspond to the feature maps at the ${i}$-th layer of {'hx6', 'hx5d', 'hx4d', 'hx3d', 'hx2d', 'hx1d'}, with ${N}$ denoting the layer quantity. Additionally, we define ${{\ell _{pd}}}$ and ${{\ell _{ps}}}$ as the distillation loss and self-study loss as outlined in {\color{blue} (\ref{eq2})} and {\color{blue} (\ref{eq3})}.

\begin{equation}
\label{eq1}
{\ell _{pf}} = \sum\limits_i^N {{{\left\| {\alpha _T^i - \alpha _S^i} \right\|}_1}}
\end{equation}

\begin{equation}
\label{eq2}
{\ell _{pd}} = {\left\| {{{\cal M}_{TAB}} - {{\cal M}_{SAB}}} \right\|_1}
\end{equation}

\begin{equation}
\label{eq3}
{\ell _{ps}} = {\left\| {{{\cal M}_{SB}} - {{\cal M}_B}} \right\|_1}
\end{equation}

\renewcommand{\algorithmicrequire}{\textbf{Input:}}
\begin{algorithm}[h]
\caption{The teaching procedure of reasoning transfer parsing.}
\label{algorithm1}
\begin{algorithmic}[1]

\REQUIRE{
pre-trained ${{{\cal T}_p}}$;${{{\cal P}_A}}$;${{{\cal P}_B}}$;${{{\cal M}_A}}$;${{{\cal M}_B}}$;${{\cal M}_B^c}$;
epoch intervals ${{e_0},{e_1},{e_2},{e_3}}$;\newline
hyper parameters ${{\lambda _1}}$,${{\lambda _2}}$,${{\lambda _3}}$;
}
\STATE ${{{\cal T}_p}}$ reasons ${{{\cal M}_{TAB}},\left\{ {\alpha _T^i} \right\}_{i = 1}^{i = 6} = {{\cal T}_p}\left( {{\cal M}_B^c,{{\cal P}_A},{\cal M}_A^{hr}} \right)}$;
\STATE ${{{\cal S}_p}}$ reasons ${{{\cal M}_{SAB}},\left\{ {\alpha _S^i} \right\}_{i = 1}^{i = 6} = {{\cal S}_p}\left( {{\cal M}_B^{hc},{{\cal P}_A},{\cal M}_A^{hr}} \right)}$;

\FOR {${e = 1 \cdots {e_3}}$}
\IF {${e < {e_1}}$}
\STATE calculate ${{\ell _{pf}}}$${\left( {N = 2} \right)}$ and ${{\ell _p} = {\lambda _1}{\ell _{pf}}}$;

\ELSIF{${{e_1} < e < {e_2}}$}
\STATE calculate ${{\ell _{pf}}}$${\left( {N = 6} \right)}$ and ${{\ell _p} = {\lambda _1}{\ell _{pf}}}$;

\ELSIF{${{e_2} < e < {e_3}}$}
\STATE calculate ${{\ell _{pf}}}$${\left( {N = 6} \right)}$, ${{\ell _{pd}}}$ and ${{\ell _p} = {\lambda _1}{\ell _{pf}} + {\lambda _2}{\ell _{pd}}}$;

\ELSE
\STATE calculate ${{\ell _{pf}}}$${\left( {N = 6} \right)}$, ${{\ell _{pd}}}$, ${{\ell _{ps}}}$ and ${{\ell _p} = {\lambda _1}{\ell _{pf}} + {\lambda _2}{\ell _{pd}} + {\lambda _3}{\ell _{ps}}}$;

\ENDIF
\ENDFOR
\STATE backward ${{\ell _p}}$ and optimize ${{{\cal S}_p}}$;
\end{algorithmic}
\end{algorithm}

\subsection{Garment Warping Teaching}
\label{sec3.3}

\subsubsection{Garment Warping Insight}
\label{sec3.3.1}

After reasoning the transfer parsing ${{{\cal M}_{SAB}}}$, we exploit it as a shape prior and guide the garment warping, which enables the garment ${{\cal I}_B^{hc}}$ of person ${B}$ to conform to the target shape ${{\cal M}_{SAB}^{hc}}$. To achieve this, there are three dominant differentiable warping tools: STN {\color{blue}\citep{ jaderberg2015spatial}}, TPS {\color{blue}\citep{ bookstein1989principal}}, and optical flow {\color{blue}\citep{ dosovitskiy2015flownet,fang2023A}}. They learn affine transformation, coarse-grained grid movement, and pixel displacement, respectively, with varying degrees of freedom. However, for garment transfer, the warping tool is required to adapt the shapes between the garment image and the target mask while preserving the style and texture of the garment. Thus, the fine-grained optical flow is the most suitable tool for this task.

Although we derive the target shape for garment warping, the ground truth of garment warping ${{\cal I}_{AB}^{hc}}$ remains unavailable. Supervising the flow warping model for garment transfer only with shape information may lead to distortion and information loss of the garment. Therefore, the content information is crucial to supervising high-DOF optical flow estimation. As paired data ${\left( {{\cal I}_A^c,{\cal I}_A^{hc}} \right)}$ is available, we can train the flow warping model for virtual try-on in a self-supervised manner, and we supervise the flow warping model for garment transfer with both shape and content information by distilling knowledge from the flow warping model for virtual try-on.

\begin{figure}[!ht]
\centering
\includegraphics[width=5.2in]{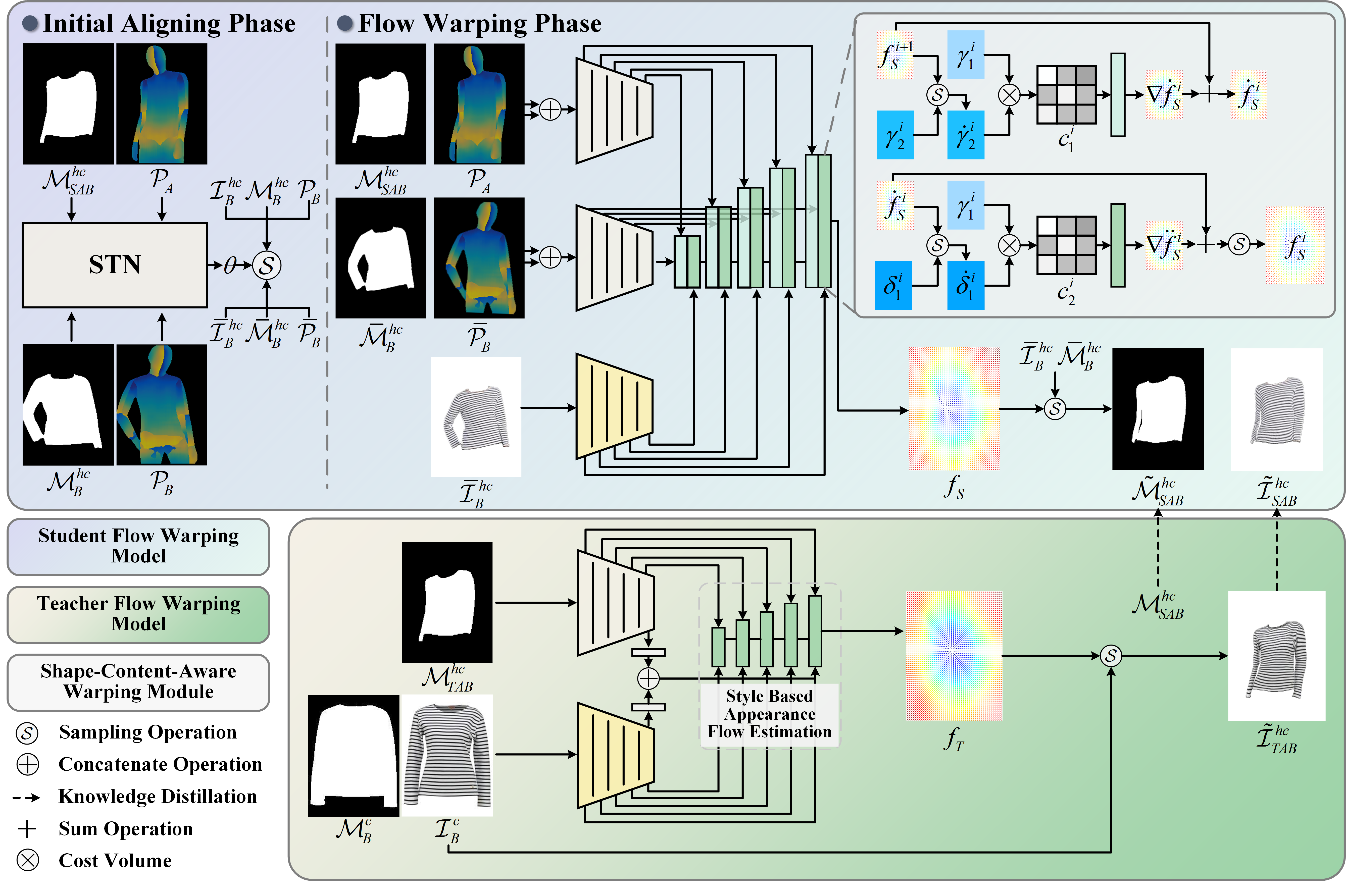}
\caption{Flow warping teaching. The student flow warping model comprises an initial aligning phase and a flow warping phase. The initial aligning phase learns an affine transformation to facilitate the position mapping, while the flow warping phase focuses on estimating a fine-grained flow to enable precise matching of the transferred garment with the target shape. The training of the student flow warping model is supervised by the shape information ${{\cal M}_{SAB}^{hc}}$ and the distillated content information ${\tilde {\cal I}_{TAB}^{hc}}$.}
\label{fig5}
\end{figure}

Specifically, as {\color{blue} Fig. \ref{fig5}} shows, we leverage the parser-based version of FSV {\color{blue}\citep{ he2022style}} as the teacher flow warping model ${{{\cal T}_w}}$, which is capable of estimating global style-based flow and refining flow locally. During its training phase, we represent the human and garment with ${{\cal M}_A^{hc}}$ and ${\left( {{\cal M}_A^c,{\cal I}_A^c} \right)}$, and exploit the ground truth ${{\cal I}_A^{hc}}$ to supervise the warping result ${\tilde {\cal I}_A^{hc}}$. Once ${{{\cal T}_w}}$ is trained, we exploit it to warp garment-in-shop ${B}$ onto person ${A}$. i.e. ${\tilde {\cal I}_{TAB}^{hc} = {{\cal T}_w}\left( {{\cal M}_{TAB}^{hc},{\cal M}_B^c,{\cal I}_B^c} \right)}$. With the supervision information ${{\cal M}_{SAB}^{hc},\tilde {\cal I}_{TAB}^{hc}}$ at the shape and content levels, we delicately devise a student flow warping model for garment transfer, inspired by the progressive flow estimation technique in {\color{blue}\citep{ ge2021parser}}, which has the initial aligning phase and the flow warping phase.

\subsubsection{Initial Aligning Phase}
\label{sec3.3.2}

\begin{figure}[!ht]
\centering
\includegraphics[width=3.7in]{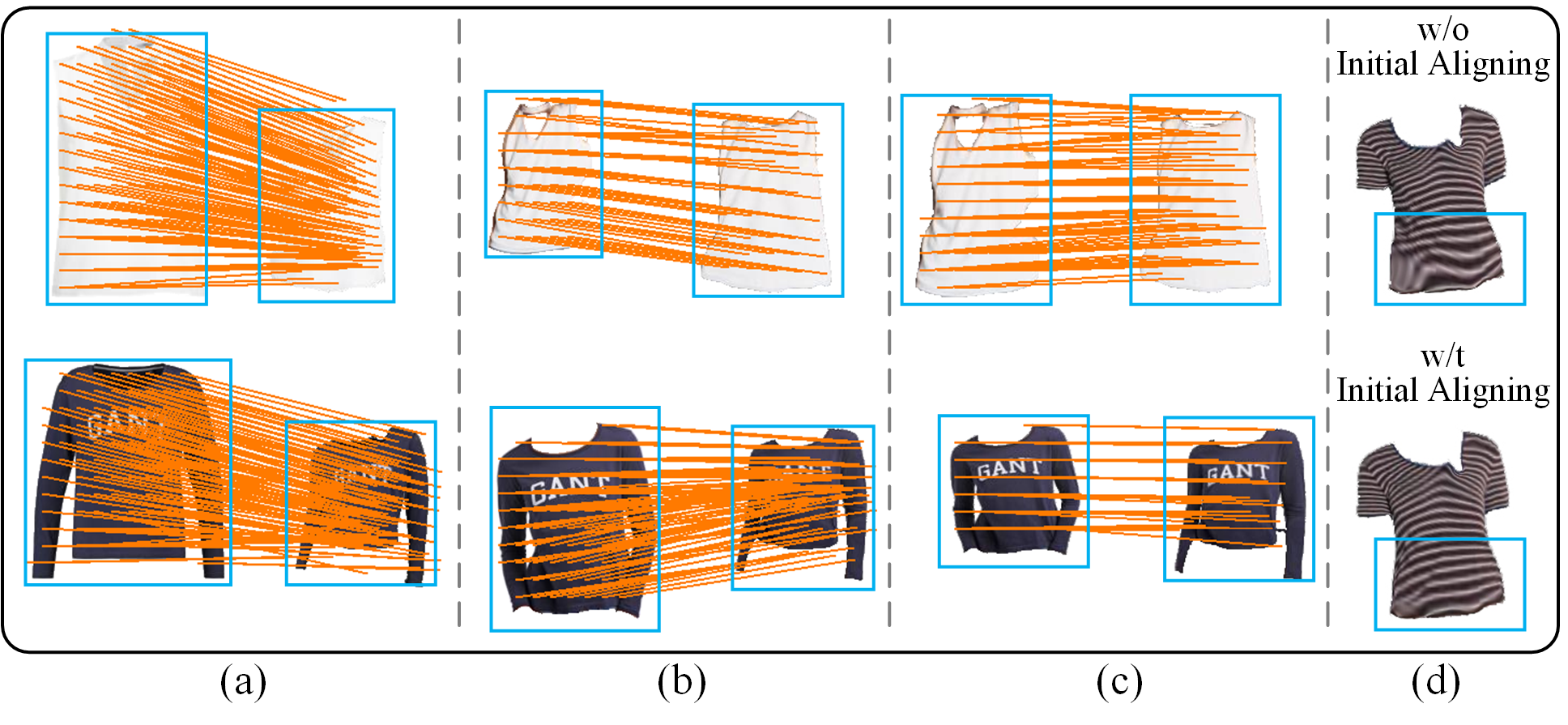}
\caption{The position mapping and shape adjustment tasks. (a) virtual try-on; (b) garment transfer without initial aligning; (c) garment transfer after initial aligning; (d) shape adjustment comparison.}
\label{fig6}
\end{figure}

We divide the garment warping problem into two subtasks, namely position mapping and shape adjustment. The former task modifies the global position of the garment by scaling, translation, and rotation, while the latter task adjusts the local shape to match the target shape precisely by non-rigid warping. For virtual try-on applications, the garment-in-shop usually has a standard pose and a high pixel ratio, as shown in {\color{blue} Fig. \ref{fig6}(a)}, which enables the flow warping model to perform both position mapping and shape adjustment tasks. However, for garment transfer, as shown in {\color{blue} Fig. \ref{fig6}(b)}, the diversity of cropping and pose of the personal image causes dramatic variation in position and shape. To address this issue, we propose to utilize STN to perform position mapping independently, as shown in {\color{blue} Fig. \ref{fig6}(c)}, which reduces the learning burden of the subsequent flow warping task. Furthermore, due to the locality of CNN, the flow warping model is not able to capture the long-range shape dependence, which may lead to distortion during precise shape matching, as shown in {\color{blue} Fig. \ref{fig6}(d)}. However, the initial aligning phase can mitigate this issue by applying the global affine transformation.

Though the paired shape representation ${\left( {{\cal M}_{SAB}^{hc},{\cal M}_B^{hc}} \right)}$ is available for the initial aligning, the binary shape representation lacks semantic information, we rely on the dense poses to provide the weak semantic information. Specifically, we feed ${{\cal M}_{SAB}^{hc},{{\cal P}_A},{\cal M}_B^{hc},{{\cal P}_B}}$ into STN to obtain an affine transformation matrix ${\theta  \in {{ \mathbb{R} } ^{3 \times 3}}}$, which is then applied to the transferred garment representation ${{\cal I}_B^{hc},{\cal M}_B^{hc},{{\cal P}_B}}$ to obtain the aligned transferred garment representation ${\bar {\cal I}_B^{hc},\bar {\cal M}_B^{hc},{\bar {\cal P}_B}}$.

\subsubsection{Flow Warping Phase}
\label{sec3.3.3}

After the initial aligning, the flow warping phase learns correspondence from coarse to fine levels and estimates a refined flow to warp garment ${\bar {\cal I}_B^{hc}}$ to target shape ${{\cal M}_{SAB}^{hc}}$ precisely. To achieve this, the shared encoder ${{E_\gamma }}$ extracts the pyramid feature maps ${\left\{ {\gamma _1^i} \right\}_{i = 1}^{i = 5}}$ and ${\left\{ {\gamma _2^i} \right\}_{i = 1}^{i = 5}}$ from weak-sematic shape representation ${\left\{ {{\cal M}_{SAB}^{hc},{{\cal P}_A}} \right\}}$ and ${\left\{ {\bar {\cal M}_B^{hc},{{\bar {\cal P}}_B}} \right\}}$ at five downsampling layers. The encoder ${{E_\delta }}$ has the same structure as ${{E_\gamma }}$ but does not share weights, which extracts the pyramid feature maps ${\left\{ {\delta _1^i} \right\}_{i = 1}^{i = 5}}$ from ${\bar {\cal I}_B^{hc}}$. Consequently, we obtain the paired shape feature maps ${\left\{ {\gamma _1^i,\gamma _2^i} \right\}}$ and the paired content-shape feature maps ${\left\{ {\gamma _1^i,\delta _1^i} \right\}}$, which serve as the condition of the subsequent shape-content-aware warping module (SCAW).

Five SCAWs learn the correspondence between feature maps at the shape level and content level. Each SCAW consists of a shape-aware stage and a content-aware stage. For convenience, we index the downsampling layers of encoders from ${i = 1}$ to ${i = 5}$ and index SCAW from ${i = 5}$ to ${i = 1}$. Consequently, the ${i}$-th SCAW has the same scale as the ${i}$-th downsampling layer of encoders and is conditioned on the last flow ${f_S^{i + 1}}$, ${\gamma _1^i}$, ${\gamma _2^i}$, and ${\delta _1^i}$. At the shape-aware stage, we first warp ${\gamma _2^i}$ to ${\dot \gamma _2^i}$ via ${f_S^{i + 1}}$ and calculate the shape cost volume ${c_1^i}$ using ${\dot \gamma _2^i}$ and ${\gamma _1^i}$ as described in {\color{blue}\citep{ilg2017flownet}}. We then estimate the shape flow residual ${\nabla \dot f_S^i}$ using four combinations \{Conv2d+LeakyRelu\} and add it to ${f_S^i}$ to obtain the shape-aware flow ${\dot f_S^i}$. At the content-aware stage, we similarly warp ${\delta _1^i}$ to ${\dot \delta _1^i}$ via ${\dot f_S^i}$ and calculate the content cost volume ${c_2^i}$ using ${\dot \delta _1^i}$ and ${\gamma _1^i}$. The content flow residual ${\nabla \ddot f_S^i}$ is estimated using another four combinations \{Conv2d+LeakyRelu\} and added to ${\dot f_S^i}$ to obtain the upsampled output ${f_S^i}$. Moreover, the first SCAW module initializes the flow by calculating ${c_1^i}$ and ${c_2^i}$ from ${\left\{ {\gamma _1^i,\gamma _2^i} \right\}}$ and ${\left\{ {\gamma _1^i,\delta _1^i} \right\}}$.

By this means, the flow is revised from coarse to fine progressively, and the pixel-level ${{f_S}}$ is estimated at the last SCAW. Finally, ${{f_S}}$ warps ${\bar {\cal I}_B^{hc}}$, ${\bar {\cal M}_B^{hc}}$ to ${\tilde {\cal I}_{SAB}^{hc}}$, ${\tilde {\cal M}_{SAB}^{hc}}$.

\subsubsection{Objective Function}
\label{sec3.3.4}
During training the student flow warping model, we formulate its objective function as {\color{blue} (\ref{eq4})}, where ${{\ell _w}}$ consists of the initial aligning loss ${{\ell _{wi}}}$, shape content loss ${{\ell _{ws}}}$, and regularization loss ${{\ell _{wr}}}$.

\begin{equation}
\label{eq4}
{\ell _w} = {\ell _{wi}} + {\ell _{ws}} + {\ell _{wr}}
\end{equation}

${{\ell _{wi}}}$ supervises the global shape matching in the initial aligning phase, consists of two items, and is defined as {\color{blue} (\ref{eq5})}. The former item punishes the shape error between ${\bar {\cal M}_B^{hc}}$ and ${{\cal M}_{SAB}^{hc}}$, whereas the latter encourages the garment pixel ratio of ${\bar {\cal M}_B^{hc}}$ to be similar to ${{\cal M}_{SAB}^{hc}}$, where ${H}$ and ${W}$ represent the height and width of masks; ${{\lambda _4}}$ and ${{\lambda _5}}$ are weights.

\begin{equation}
\label{eq5}
{\ell _{wi}} = {\lambda _4}{\left\| {\bar {\cal M}_B^{hc} - {\cal M}_{SAB}^{hc}} \right\|_1} + {\lambda _5}{\left\| {\sum\limits_{x,y}^{H,W} {\bar {\cal M}_B^{hc}}  - \sum\limits_{x,y}^{H,W} {{\cal M}_{SAB}^{hc}} } \right\|_1}
\end{equation}

${{\ell _{ws}}}$ supervises the knowledge distillation from ${{{\cal T}_w}}$ to ${{{\cal S}_w}}$, which consists of two items as {\color{blue} (\ref{eq6})}. The first item punishes the shape error between ${\tilde {\cal M}_{SAB}^{hc}}$ and ${{\cal M}_{SAB}^{hc}}$, while the second and third items punish the image difference between ${\tilde {\cal I}_{TAB}^{hc}}$ and ${\tilde {\cal I}_{SAB}^{hc}}$ at both the pixel and perceptual levels, where ${\phi \left(  \cdot  \right)}$ is the VGG loss function. We adopt the pre-trained VGG-19 model {\color{blue}\citep{simonyan2014very}} to calculate ${\phi \left(  \cdot  \right)}$ at the 'relu1${\_}$1', 'relu2${\_}$1', 'relu3${\_}1$', and 'relu4${\_}1$' layers. ${{\lambda _6}}$,${{\lambda _7}}$,${{\lambda _8}}$ are weights.

\begin{equation}
\label{eq6}
{\ell _{ws}} = {\lambda _6}{\left\| {\tilde {\cal M}_{SAB}^{hc} - {\cal M}_{SAB}^{hc}} \right\|_1} + {\lambda _7}{\left\| {\tilde {\cal I}_{SAB}^{hc} - \tilde {\cal I}_{TAB}^{hc}} \right\|_1} + {\lambda _8}{\left\| {\sum\limits_{i = 0}^4 {\phi \left( {\tilde {\cal I}_{SAB}^{hc}} \right) - \phi \left( {\tilde {\cal I}_{TAB}^{hc}} \right)} } \right\|_1}
\end{equation}

${{\ell _{wr}}}$ is the penalty term of the flow displacement, which consists of two items as {\color{blue} (\ref{eq7})}. We exploit both the one-order constraint in {\color{blue}\citep{he2022style}} and the second-order constraint in {\color{blue}\citep{ge2021parser}} to be ${{\ell _{wr1}}}$ and ${{\ell _{wr2}}}$. ${{\lambda _9}}$ and ${{\lambda _{10}}}$ are weights.

\begin{equation}
\label{eq7}
{\ell _{wr}} = {\lambda _9}{\ell _{wr1}} + {\lambda _{10}}{\ell _{wr2}}
\end{equation}

\subsection{Arm Regrowth}
\label{sec3.4}
The upper body skin category has two cases in garment transfer: one case is to transfer a short-sleeve garment onto a person who wears a long-sleeve garment, and another case is to transfer a long-sleeve garment onto a person who wears a short-sleeve garment. To distinguish these two cases precisely, we compare the source and target shapes of upper body skin category, i.e. ${{\cal M}_A^{ha}}$ and ${{\cal M}_{SAB}^{ha}}$. The former case is defined as ${{\cal M}_A^{ha} > \left( {{\cal M}_{SAB}^{ha} \cup {\cal M}_A^{ha}} \right)}$ while the latter case is defined as ${{\cal M}_A^{ha} < \left( {{\cal M}_{SAB}^{ha} \cup {\cal M}_A^{ha}} \right)}$. For the former case, we just inherit the part of existing skin. For the latter case, we need to design a model to infer new exposure skin, and name this task ‘arm regrowth’.

\begin{figure}[!ht]
\centering
\includegraphics[width=4.2in]{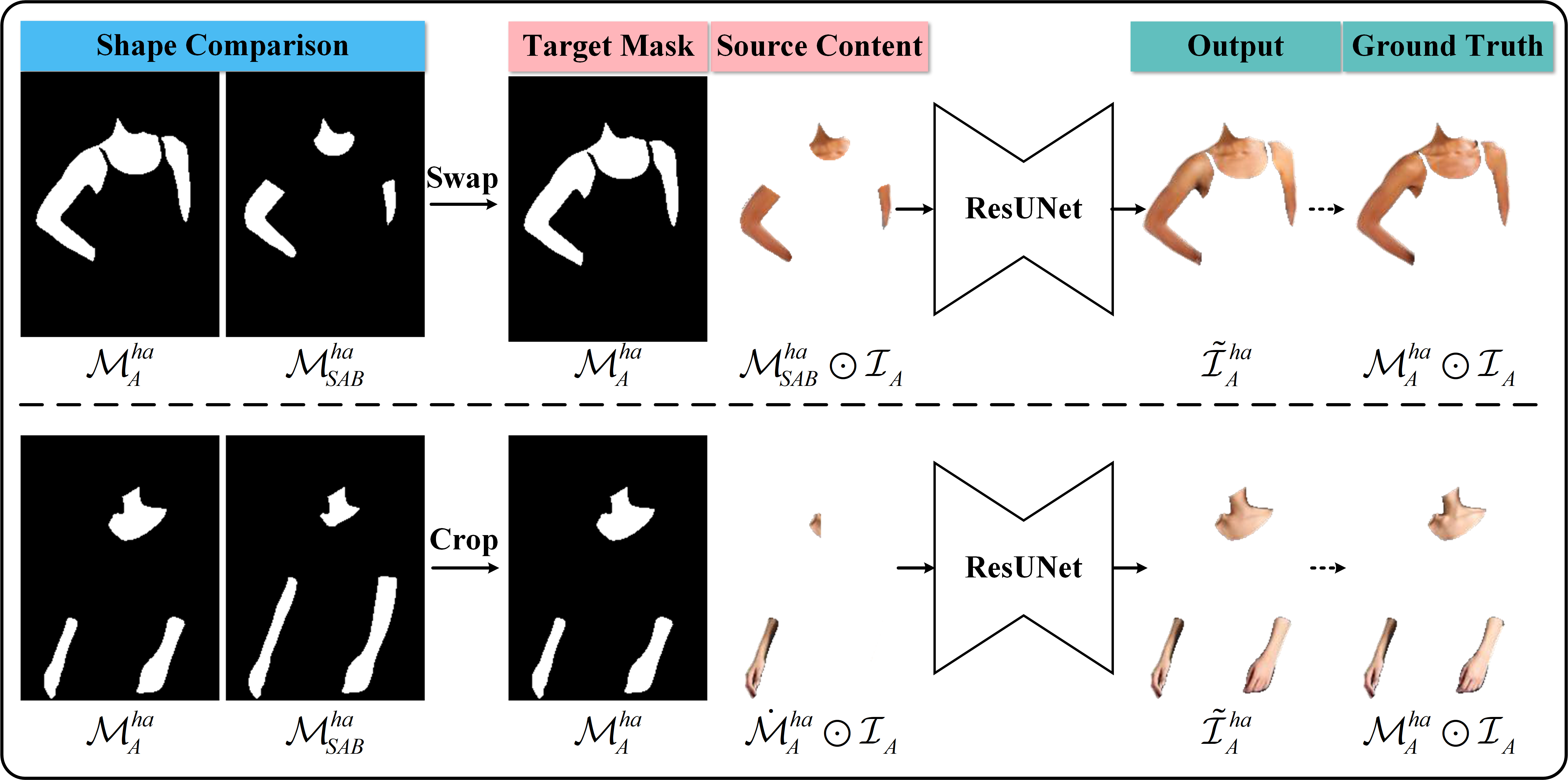}
\caption{The training phase of ${{G_r}}$.}
\label{fig7}
\end{figure}

As shown in {\color{blue} Fig. \ref{fig7}}, we train the arm regrowth model ${{G_r}}$ in a self-supervision manner, where ${{G_r}}$ is implemented by a ResUnet {\color{blue}\citep{ zhang2018road}}. In the training phase, we adjust both two cases into an arm regrowth task with a ground truth. For the former case, we swap inputs, take ${{\cal M}_{SAB}^{ha} \odot {\cal I}_A^{ha}}$ and ${{\cal M}_A^{ha}}$ as the source content and target shape, and leverage ${{\cal M}_A^{ha} \odot {\cal I}_A^{}}$ as the ground truth. ${{G_r}}$ predicts ${\tilde {\cal I}_A^{ha}}$ to simulate the arm regrowth task as ${\tilde {\cal I}_A^{ha} = {G_r}\left( {{\cal M}_{SAB}^{ha} \odot {\cal I}_A^{ha},{\cal M}_A^{ha}} \right)}$. For the latter case, though inputs conform to the arm regrowth task, the ground truth is lacking. Therefore, we randomly crop ${{\cal M}_A^{ha}}$ to ${\dot {\cal M}_A^{ha}}$, take ${\dot {\cal M}_A^{ha} \odot {{\cal I}_A}}$ and ${{\cal M}_A^{ha}}$ as the source content and target shape, and utilize ${{\cal M}_A^{ha} \odot {\cal I}_A^{}}$ as the ground truth. Then, ${{G_r}}$ predicts ${\tilde {\cal I}_A^{ha}}$ as ${\tilde {\cal I}_A^{ha} = {G_r}\left( {\dot {\cal M}_A^{ha} \odot {{\cal I}_A},{\cal M}_A^{ha}} \right)}$. Moreover, the cropping strategy is to remove half-plane shape on the horizontal or vertical axis, and these horizontal or vertical probability is equal.

In the testing phase, for the former case, we multiply ${{\cal M}_{SAB}^{ha}}$ with ${{\cal M}_A^{ha} \odot {{\cal I}_A}}$ to form ${{\cal I}_{SAB}^{ha}}$, and for the latter case, we use pre-trained ${{G_r}}$ to predict ${{\cal I}_{SAB}^{ha}}$ via ${\tilde {\cal I}_{SAB}^{ha} = {G_r}\left( {{\cal I}_A^{ha},{\cal M}_{SAB}^{ha}} \right)}$.

We design its objective function ${{\ell _r}}$  as {\color{blue} (\ref{eq8})} where ${{\ell _{r1}}}$,${{\ell _{rc}}}$ are the L1 and content losses; ${{\lambda _{11}}}$,${{\lambda _{12}}}$ are weights. ${{\ell _{r1}}}$ and ${{\ell _{rc}}}$ are defined as {\color{blue} (\ref{eq9})} and {\color{blue} (\ref{eq10})} where ${\phi \left(  \cdot  \right)}$ is the same as {\color{blue} (\ref{eq6})}.

\begin{equation}
\label{eq8}
{\ell _r} = {\lambda _{11}}{\ell _{r1}} + {\lambda _{12}}{\ell _{rc}}
\end{equation}

\begin{equation}
\label{eq9}
{\ell _{r1}} = {\left\| {\tilde {\cal I}_A^{ha} - {\cal M}_A^{ha} \odot {{\cal I}_A}} \right\|_1}
\end{equation}

\begin{equation}
\label{eq10}
{\ell _{rc}} = {\left\| {\sum\limits_{i = 1}^4 {\phi \left( {\tilde {\cal I}_A^{ha}} \right)}  - \sum\limits_{i = 1}^4 {\phi \left( {{\cal M}_A^{ha} \odot {{\cal I}_A}} \right)} } \right\|_1}
\end{equation}

\section{Experiment}
\label{sec4}

\subsection{Implementation Details}
\label{sec4.1}

We exploit the Zalando dataset {\color{blue}\citep{han2018viton}} to train and evaluate our proposed method\footnote{We have gotten the using permission for the research purpose from its author Xintong Han.}. This dataset is further cleaned so that the training set and test set consist of 11565 and 1698 paired data with a resolution of ${256 \times 192}$. We randomly match person ${A}$ and ${B}$ to form the garment transfer pair. All experiments were conducted on a single 3090 NVIDIA GPU by Adam optimizer. We present the hyperparameters of garment transfer models in the training phase in {\color{blue} Table \ref{table1}} and provide the model complexity in {\color{blue} Table \ref{table2}}.

\begin{table}[!ht]
\caption{The hyperparameters in training garment transfer models.}
\label{table1}
\centering
\resizebox{\columnwidth}{!}{
\begin{tabular}{ccccc}
\toprule Network    &  LR & Iterations & Batch & Weights   \\ \midrule
                     ${{{\cal S}_p}}$  &  ${2 \times {e^{ - 4}}}$  & 18.1 K  & 16 & \makecell[l]{${{\lambda _1} = 0.1}$, ${{\lambda _2} = 60.0}$, ${{\lambda _3} = 60.0}$}\\
                     ${{{\cal S}_w}}$  &  ${1 \times {e^{ - 4}}}$  & 72.2 K  & 8  &  \makecell[l]{${{\lambda _4} = 0.6}$, ${{\lambda _5} = 6 \times {e^{ - 3}}}$,
                     ${{\lambda _6} = 2.0}$, \\${{\lambda _7} = 1.0}$, ${{\lambda _8} = 0.2}$, ${{\lambda _9} = 6.0}$, ${{\lambda _{10}} = 1 \times {e^{ - 2}}}$} \\
                     ${{G_r}}$         &  ${1 \times {e^{ - 4}}}$  & 18.1 K  &16 & \makecell[l]{${{\lambda _{11}} = 10.0}$,${{\lambda _{12}} = 0.2}$}\\ \bottomrule

\end{tabular}
}
\end{table}

\begin{table}[!ht]
\caption{The model complexity.}
\label{table2}
\centering
\begin{tabular}{ccccc}
\toprule Model    &  ${{{\cal S}_p}}$ & ${{{\cal S}_w}}$ & ${{G_r}}$ & All   \\ \midrule
          Params/M&  44.10&	26.66 &	43.90 &	114.66\\
          FlOPs/G &28.62  &	55.05 &	21.86 &	105.53 \\\bottomrule
\end{tabular}
\end{table}

\subsection{Ablation Study}
\label{sec4.2}

\subsubsection{Transfer Parsing Reasoning}
\label{sec4.2.1}
\textbf{Metrics} In this section, we conduct the ablation study of transfer parsing reasoning to verify the necessity of knowledge distillation  in training ${{{\cal S}_p}}$, the rationality of the phase setting in the teaching procedure, and the appropriateness of the backbone and pose form. We exploit the intersection over union (IoU) and class pixel accuracy (CPA) as metrics, both of which range from 0 to 1. Since the entanglement between the garment and pose is crucial in transfer parsing reasoning, we focus on the metrics values of the garment and upper body skin categories. Thus, we compare ${{{\cal M}_{SAB}}}$ with ${{{\cal M}_{TAB}}}$ to verify the knowledge distillation effect, denoted as ${{\rm{IoU}}\left( {{\cal M}_{TAB}^{ha},{\cal M}_{SAB}^{ha}} \right)}$, ${{\rm{IoU}}\left( {{\cal M}_{TAB}^{hc},{\cal M}_{SAB}^{hc}} \right)}$, ${{\rm{CPA}}\left( {{\cal M}_{TAB}^{ha}, {\cal M}_{SAB}^{ha}} \right)}$, and ${{\rm{CPA}}\left( {{\cal M}_{TAB}^{hc},{\cal M}_{SAB}^{hc}} \right)}$. And we compare ${{{\cal M}_{SB}}}$ with ${{{\cal M}_B}}$ to verify the reasoning rationality with ground truth, denoted as ${{\rm{IoU}}\left( {{\cal M}_{SB}^{ha},{\cal M}_B^{ha}} \right)}$, ${{\rm{IoU}}\left( {{\cal M}_{SB}^{hc},{\cal M}_B^{hc}} \right)}$, ${{\rm{CPA}}\left( {{\cal M}_{SB}^{ha}, {\cal M}_B^{ha}} \right)}$, and ${{\rm{CPA}}\left( {{\cal M}_{SB}^{hc},{\cal M}_B^{hc}} \right)}$.

\begin{figure}[!ht]
\centering
\includegraphics[width=3.3in]{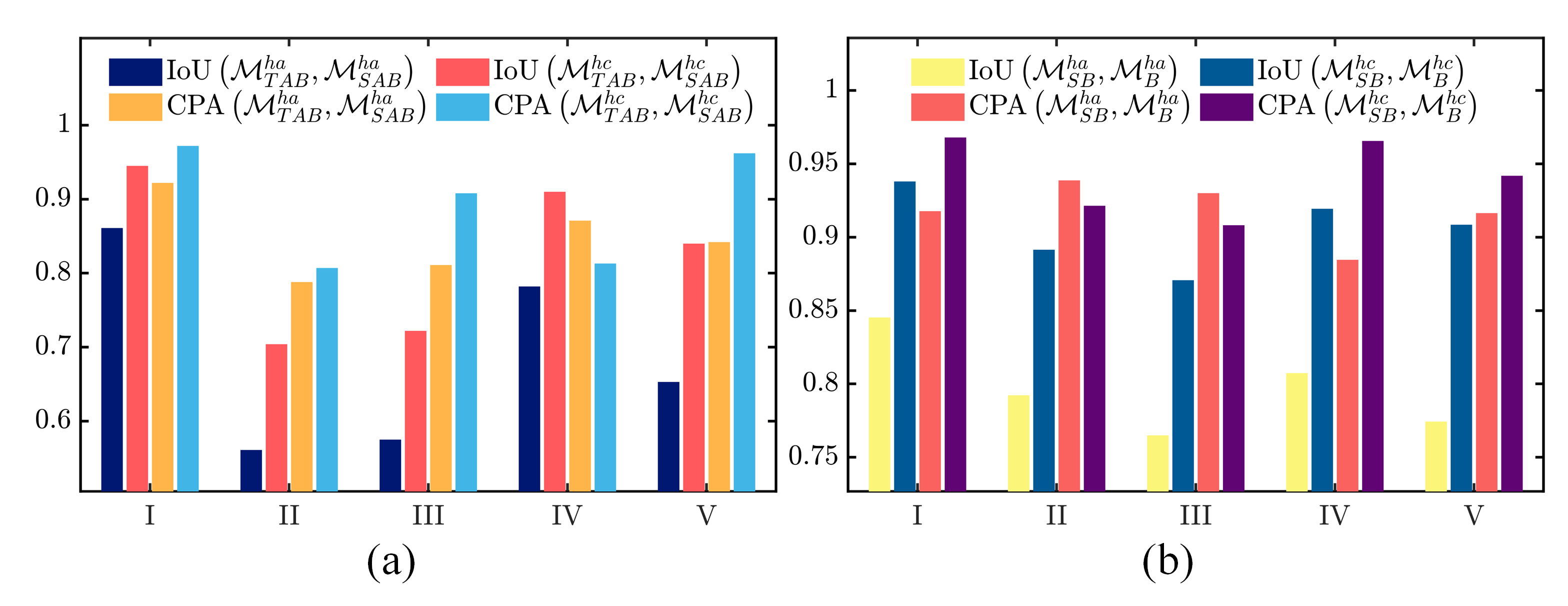}
\caption{Quantitative results in verifying the necessity of knowledge distillation. Cases I-V are the baseline; affine transformation, affine transformation and cropping; affine transformation, cropping, and flipping; and cycle consistency.}
\label{fig8}
\end{figure}

\begin{figure}[!ht]
\centering
\includegraphics[width=3in]{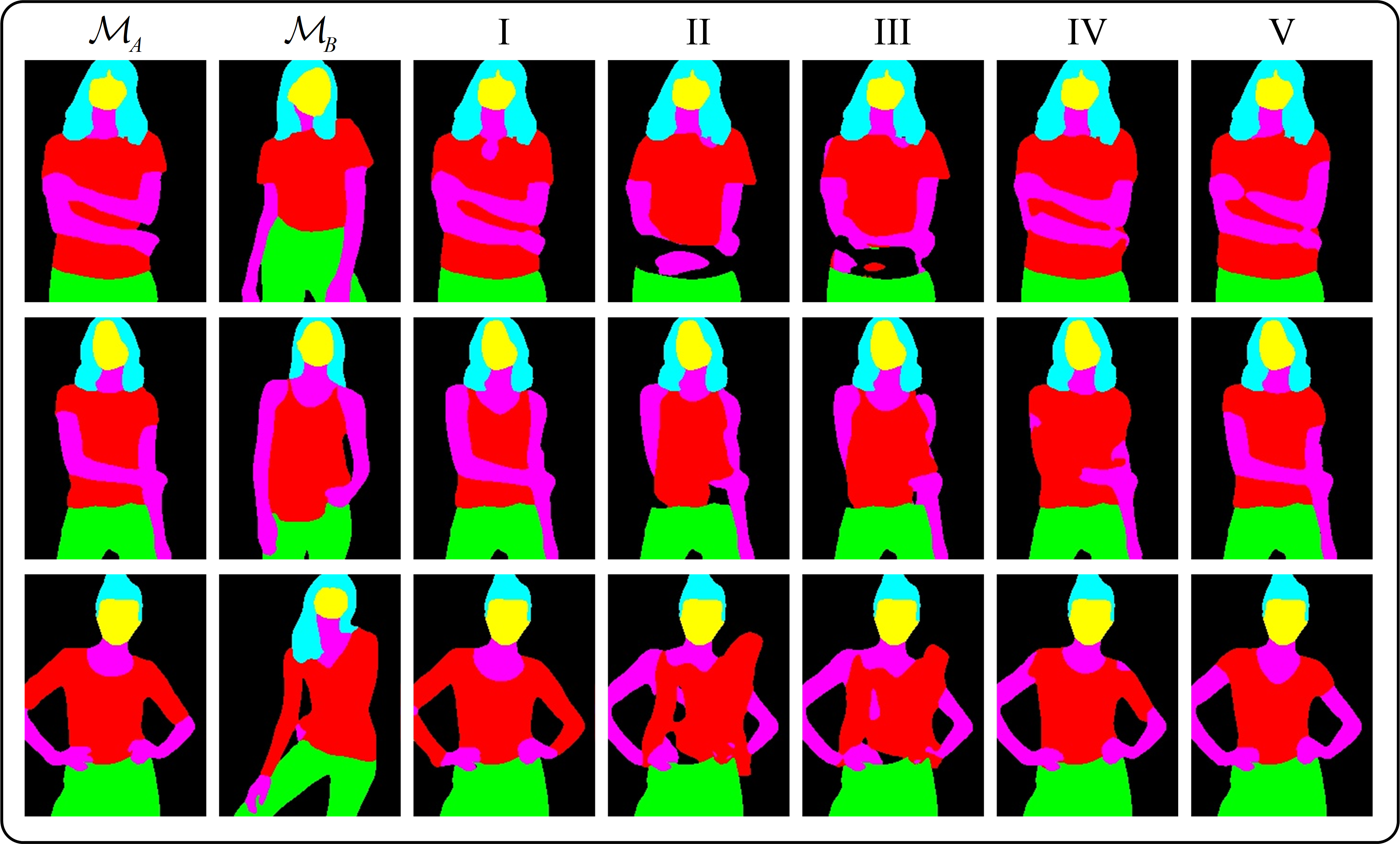}
\caption{Qualitative results in verifying the necessity of knowledge distillation. Cases I-V are the same as {\color{blue} Fig. \ref{fig8}}.}
\label{fig9}
\end{figure}

\textbf{The Necessity of Knowledge Distillation} As described in {\color{blue} Sec. \ref{sec3.2.1}}, without knowledge distillation, we train ${{{\cal S}_p}}$ in a self-supervised manner as comparisons. Specifically, during the training phase, ${{{\cal S}_p}}$ is conditioned on augmented ${{\cal M}_A^{hc}}$, ${{\cal M}_A^{hr}}$, and ${{{\cal P}_A}}$, while the ground truth ${{{\cal M}_A}}$ is leveraged to supervise the output of ${{{\cal S}_p}}$. The traditional data augmentation operations, including affine transformation, cropping, and flipping, are used in three different combinations: affine transformation; affine transformation and cropping; and affine transformation, cropping, and flipping. To further improve the performance, we adopted the cycle consistency strategy from {\color{blue}\citep{ge2021disentangled}}, which involves dressing ${{\cal M}_B^{hc}}$ onto person ${A}$ to generate ${{\tilde {\cal M}_{AB}}}$ and then dressing ${\tilde {\cal M}_{AB}^{hc}}$ onto person ${B}$ again, supervised by the ground truth ${{{\cal M}_B}}$. The quantitative and qualitative results are shown in {\color{blue} Fig. \ref{fig8}} and {\color{blue} Fig. \ref{fig9}}, respectively.

For cases II and III, the translation invariance of CNN allows ${{{\cal S}_p}}$ to learn the pose feature from ${{\cal M}_A^{hc}}$ even after affine transformation and cropping. This enables the transfer parsing reasoning to degenerate into rigid transformation learning, resulting in ${{\cal M}_B^{hc}}$ being simply pasted onto ${{\cal M}_A^{hr}}$ to form ${{{\cal M}_{SAB}}}$, leading to a reduction of 34.03\% and 24.55\% in ${{\rm{IoU}}\left( {{\cal M}_{TAB}^{ha},{\cal M}_{SAB}^{ha}} \right)}$ and ${{\rm{IoU}}\left( {{\cal M}_{TAB}^{hc},{\cal M}_{SAB}^{hc}} \right)}$. For case IV, the flipping operation goes against the CNN translation invariance, but the style and shape of the garment are completely changed. In consequence, ${{{\cal S}_p}}$ has a poor performance in capturing these features, leading to ${{{\cal M}_{SAB}}}$ having the shape and style deviation, especially when ${{\cal M}_B^{hc}}$ is in a complex pose. And ${{\rm{IoU}}\left( {{\cal M}_{TAB}^{ha},{\cal M}_{SAB}^{ha}} \right)}$, ${{\rm{IoU}}\left( {{\cal M}_{TAB}^{hc},{\cal M}_{SAB}^{hc}} \right)}$ reduce by 9.17\% and 3.49\%, respectively. For case V, the CNN model tends to take a shortcut, but cycle consistency fails to supervise the intermediate result ${{\tilde {\cal M}_{AB}}}$. Therefore, cycle consistency prefers less variation, which results in the garment style in ${{{\cal M}_{SAB}}}$ being improperly changed to a T-shirt, irrespective of the style of the transferred garment. This leads to a reduction of ${{\rm{CPA}}\left( {{\cal M}_{TAB}^{ha},{\cal M}_{SAB}^{ha}} \right)}$ and ${{\rm{CPA}}\left( {{\cal M}_{TAB}^{hc},{\cal M}_{SAB}^{hc}} \right)}$ by 9.17\% and 3.49\%, respectively.

\begin{figure}[!ht]
\centering
\includegraphics[width=4.2in]{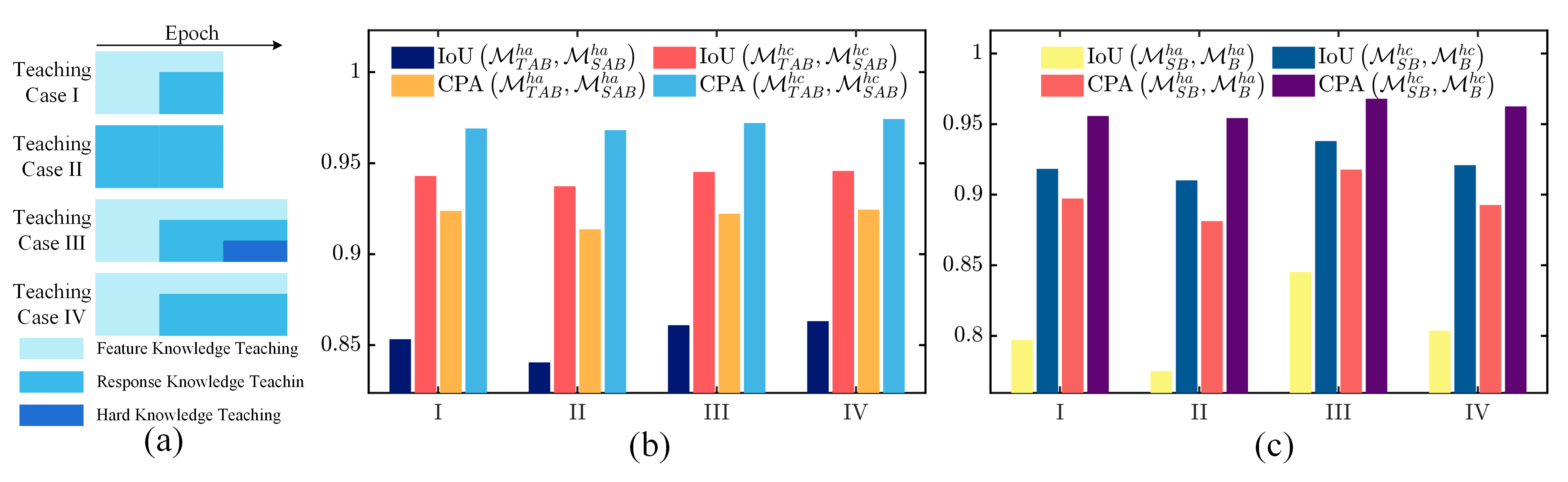}
\caption{Quantitative results in verifying the phase setting rationality. (a) four cases; (b) the teacher similarity; (c) the ground truth similarity.}
\label{fig10}
\end{figure}

\begin{figure}[!ht]
\centering
\includegraphics[width=3.2in]{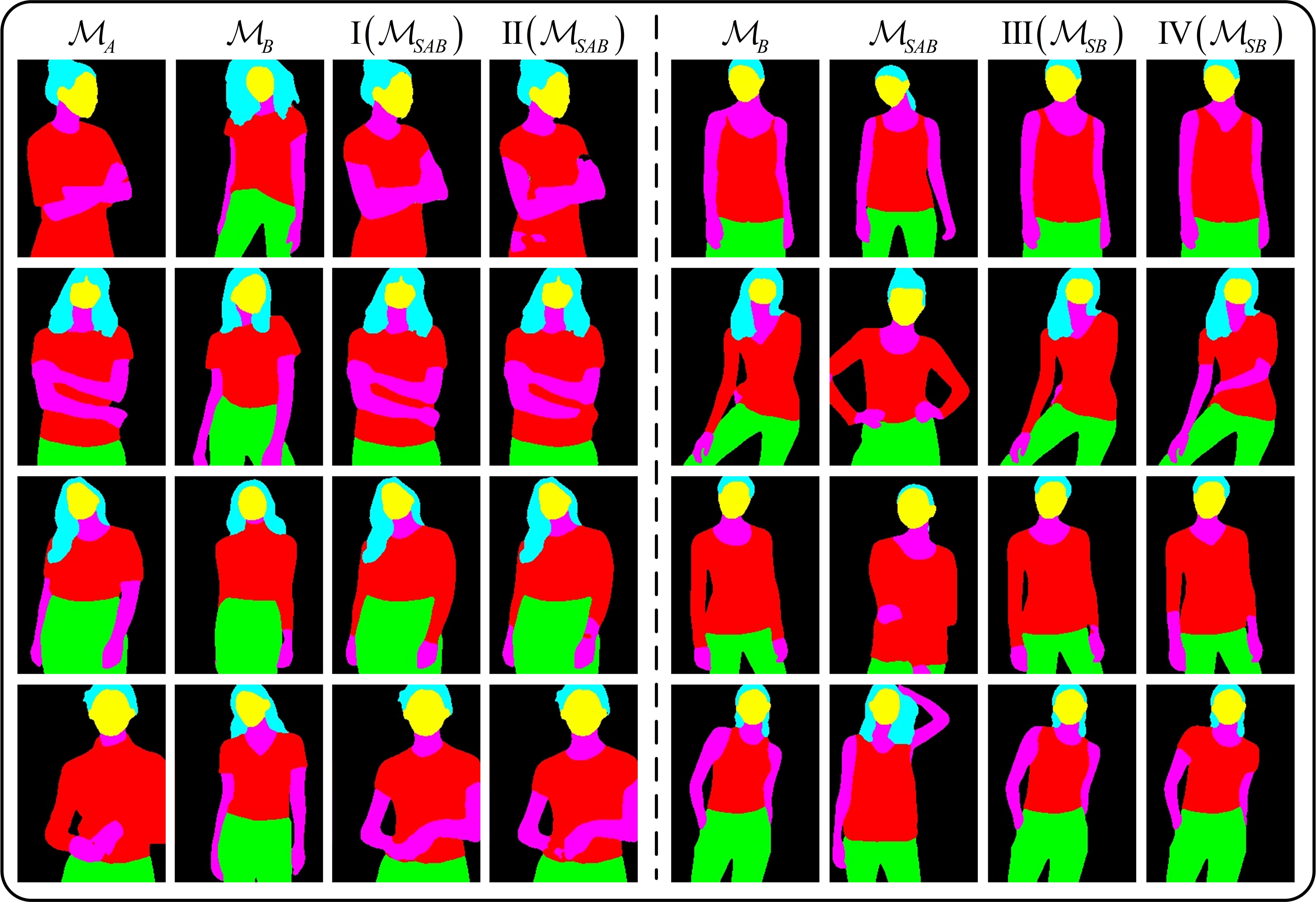}
\caption{Qualitative comparisons in verifying the phase setting rationality.}
\label{fig11}
\end{figure}

\textbf{The Rationality of Phase Setting} As described in {\color{blue} Sec. \ref{sec3.2.4}}, we design four cases to verify the rationality of phase setting. The former two cases are designated to verify the effectiveness of the feature knowledge teaching phase, while the latter two cases are intended to verify the self-study phase. As {\color{blue} Fig. \ref{fig10}(a)} shows, in case I, we follow {\color{blue} Algorithm. \ref{algorithm1}} until epoch ${{e_3}}$, while in case II, we only distill response knowledge within the epoch range of ${\left[ {0,{e_3}} \right]}$. Case III is designed to follow {\color{blue} Algorithm. \ref{algorithm1}} until epoch ${{e_4}}$, and case IV follows {\color{blue} Algorithm. \ref{algorithm1}} within the epoch range of ${\left[ {0,{e_3}} \right]}$, but only distills the feature and response knowledge during ${\left[ {{e_3},{e_4}} \right]}$. The quantitative and qualitative results are shown in {\color{blue} Fig. \ref{fig10}} and {\color{blue} Fig. \ref{fig11}}, respectively.

Compared with case I, case II does not include the reasoning process teaching, it only simulates ${{{\cal T}_p}}$ at the response level but neglects to learn the semantic feature entanglement and human structure reconstruction from ${{{\cal T}_p}}$. Consequently, case II fails to comprehend the semantic features of garment style and global human pose, leading to fragmentary arm and sleeve variations, and a 1.49\% and 1.10\% decrease in ${{\rm{IoU}}\left( {{\cal M}_{TAB}^{ha},{\cal M}_{SAB}^{ha}} \right)}$ and ${{\rm{CPA}}\left( {{\cal M}_{TAB}^{ha},{\cal M}_{SAB}^{ha}} \right)}$, respectively. To compare case III and case IV, we adopt the more challenging reasoning ${{{\cal M}_{SB}} = {{\cal S}_p}\left( {{{\cal P}_B},{\cal M}_B^{hr},{\cal M}_{SAB}^{hc}} \right)}$. Case IV simulates ${{{\cal T}_p}}$ so excessively that it is misled by wrong reasoning instances, such as comprehending a long-sleeve garment as a short-sleeve one. As a result, case IV has poor performance in reasoning ${{{\cal M}_{SB}}}$ compared to reasoning ${{{\cal M}_{SAB}}}$. In contrast, case III can revise the wrong teaching in the self-study phase and has a similar performance in reasoning ${{{\cal M}_{SB}}}$ and ${{{\cal M}_{SAB}}}$. It indicates that the reasoning of case III is closer to the ground truth rather than just following ${{{\cal T}_p}}$. Additionally, the quantitative results of case III in reasoning ${{{\cal M}_{SB}}}$ show a significant increase, indicating that the self-study phase allows ${{{\cal S}_p}}$ to be more stable by feeding more transfer instances.

\begin{figure}[!ht]
\centering
\includegraphics[width=3.8in]{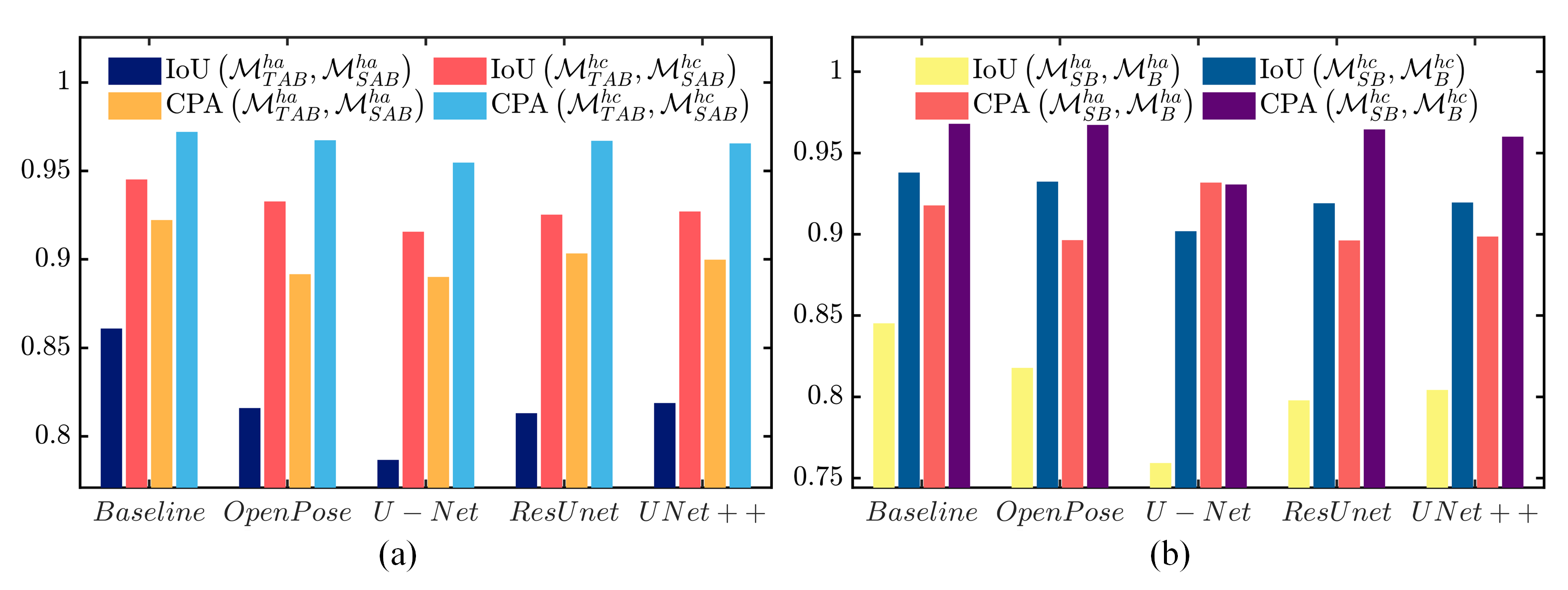}
\caption{Quantitative results in verifying the appropriateness of backbone and pose form. (a) the teacher similarity; (b) the ground truth similarity.}
\label{fig12}
\end{figure}

\begin{figure}[!ht]
\centering
\includegraphics[width=2.8in]{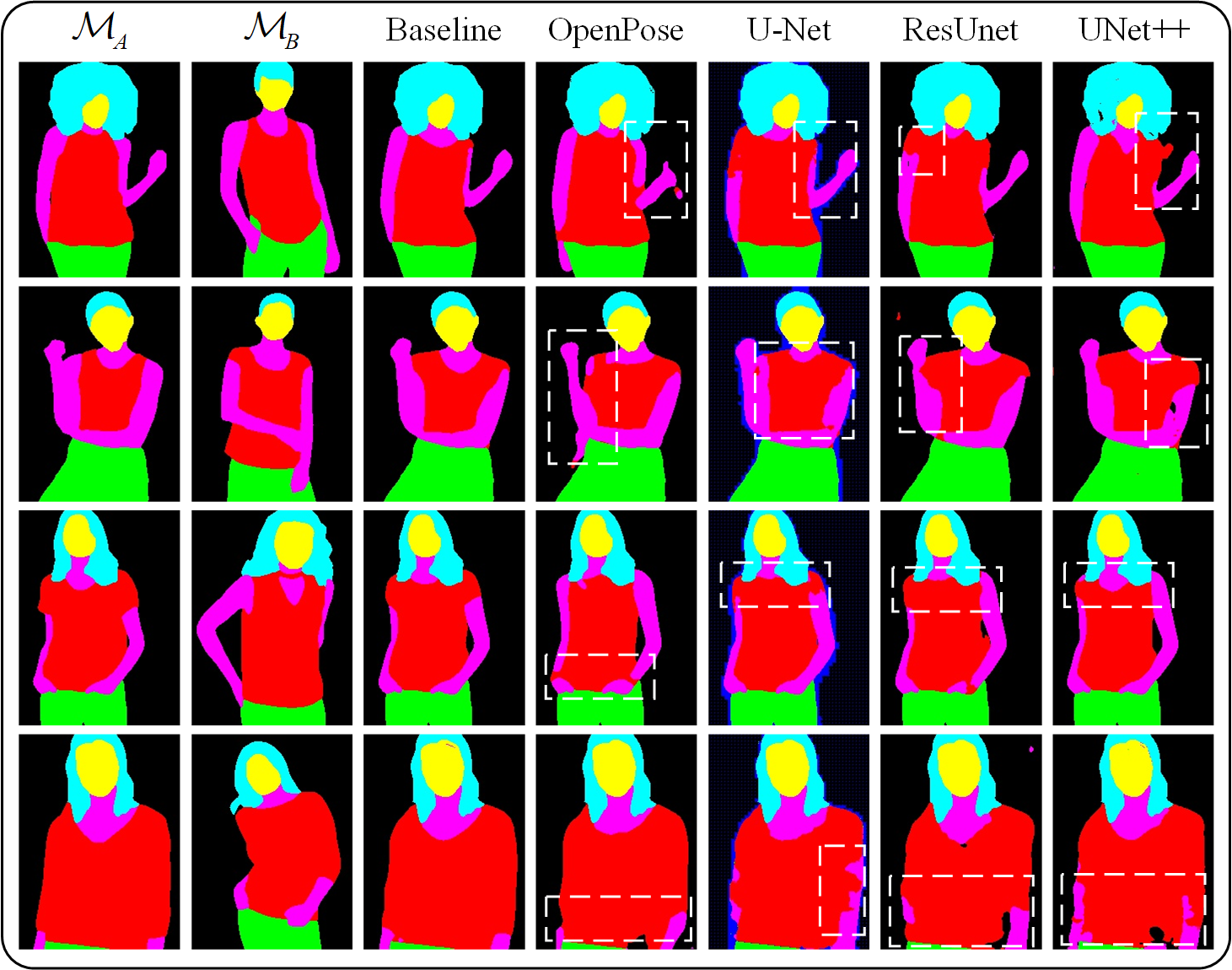}
\caption{Qualitative comparisons in verifying the appropriateness of backbone and pose form.}
\label{fig13}
\end{figure}

\textbf{The Appropriateness of Backbone and Pose Form} We compare the appropriateness of different backbones for the parsing reasoning task, including U2-Net {\color{blue}\citep{ qin2020u2}}, U-Net {\color{blue}\citep{ ronneberger2015u}}, ResUnet {\color{blue}\citep{ zhang2018road}}, and UNet++ {\color{blue}\citep{ zhou2019unet++}}. We compare the function of DensePose and OpenPose in reasoning. The quantitative and qualitative results are shown in {\color{blue} Fig. \ref{fig12}} and {\color{blue} Fig. \ref{fig13}}, respectively.

The U2-Net backbone utilizes a deep architecture by nesting U-Net and is adept at capturing semantic features of garments and poses. It excels in reconstructing and propagating local details by fusing multiscale features, resulting in better preservation of arm shape and garment details compared to other backbones, as indicated by a reduction of ${{\rm{IoU}}\left( {{\cal M}_{SB}^{hc},{\cal M}_B^{hc}} \right)}$ and ${{\rm{IoU}}\left( {{\cal M}_{SB}^{hc},{\cal M}_B^{hc}} \right)}$ by 6.87\% and 2.60\%, respectively. Thus, U2-Net is deemed more suitable for the parsing reasoning task among the considered backbones. On the other hand, OpenPose only provides information on joint positions, and thus, fails to reconstruct fine arm shapes due to the lack of arm shape prior. In contrast, DensePose provides human structure and shape prior and results in an increase of ${{\rm{IoU}}\left( {{\cal M}_{SAB}^{ha},{\cal M}_{TAB}^{ha}} \right)}$ and ${{\rm{IoU}}\left( {{\cal M}_B^{ha},{\cal M}_{SB}^{ha}} \right)}$ by 5.51\% and 3.34\%, respectively.

\subsubsection{Garment Warping}
\label{sec4.2.2}
\textbf{Metrics} In this section, we verify the necessity of knowledge distillation in training  , the appropriateness of warping method, and the rationality of phase setting. Similar to {\color{blue} Sec. \ref{sec4.2.1}}, we compare ${\tilde {\cal I}_{SAB}^{hc}}$ with ${\tilde {\cal I}_{TAB}^{hc}}$ to verify the knowledge distillation and compare ${\tilde {\cal I}_{SB}^{hc}}$ with ${{\cal I}_B^{hc}}$ to verify the garment warping authenticity to the ground truth. We exploit the low-level full-reference metrics, high-level full-reference metrics, and no-reference subjective metric. The low-level full-reference metrics consist of MSE, SSIM, and PSNR; the high-level full-reference metrics consist of FID {\color{blue}\citep{ heusel2017gans}} and LPIPS {\color{blue}\citep{ zhang2018unreasonable}}; the no-reference subjective metric is IS {\color{blue}\citep{ salimans2016improved}}.

\begin{figure}[!ht]
\centering
\includegraphics[width=3.2in]{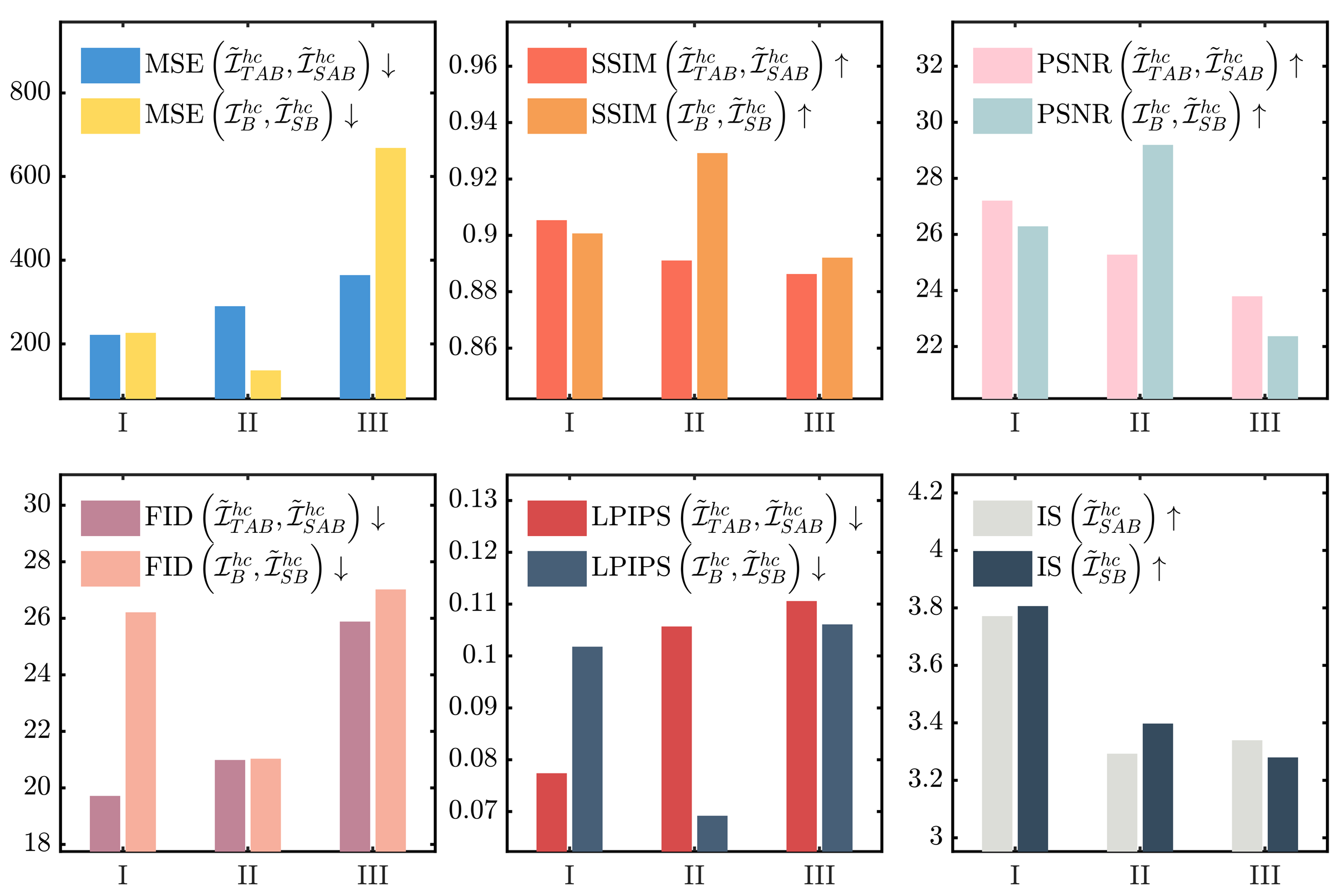}
\caption{Quantitative results in verifying the necessity of knowledge distillation. Cases I-III are baseline, supervision only with shape information, and training in cycle consistency.}
\label{fig14}
\end{figure}

\begin{figure}[!ht]
\centering
\includegraphics[width=2.2in]{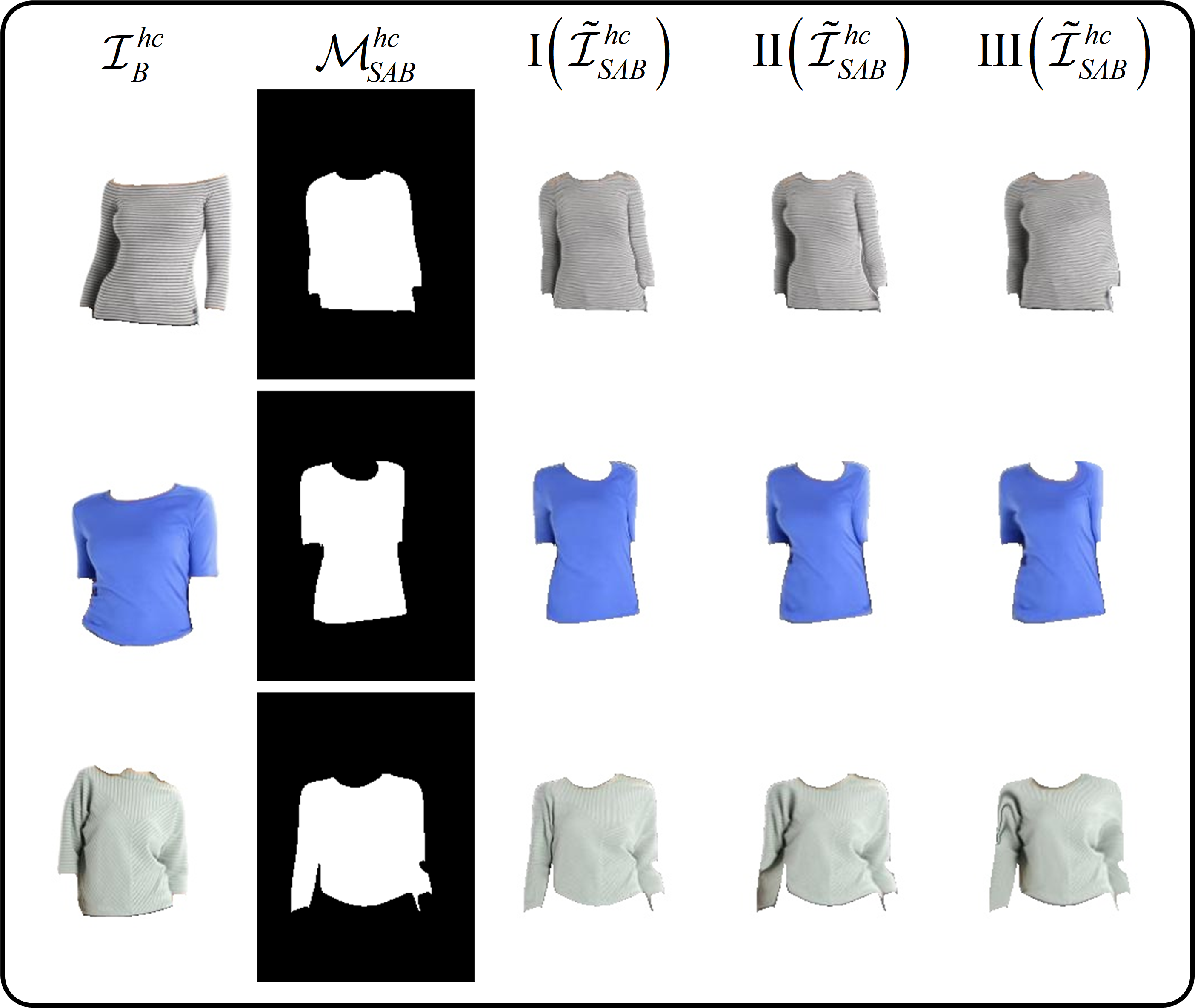}
\caption{Qualitative results in verifying the necessity of knowledge distillation.}
\label{fig15}
\end{figure}

\textbf{The Necessity of Knowledge Distillation} As outlined in {\color{blue} Sec. \ref{sec3.3.1}}, we investigate the necessity of knowledge distillation in training ${{{\cal S}_w}}$ by comparing three cases. The first case is our baseline. The second case involves training ${{{\cal S}_w}}$ solely using shape information ${{\cal M}_{SAB}^{hc}}$, without knowledge distillation from ${{{\cal T}_w}}$. The third case involves training ${{{\cal S}_w}}$ by enforcing cycle consistency, as described in {\color{blue} Sec. \ref{sec4.2.1}}. The comparison results are presented in {\color{blue} Fig. \ref{fig14}} and {\color{blue}\ref{fig15}}.

Case II only warps the boundary of ${{\cal I}_B^{hc}}$ towards ${{\cal M}_{SAB}^{hc}}$, while the internal structure of the garment remains unchanged because the shape information does not guide warping inside the garment. In contrast, in case I, the distillated content information ${\tilde {\cal I}_{TAB}^{hc}}$ supervises the warp of the internal content of the garment to reflect pose changes. This results in a more realistic warp that conforms to the global semantics of garment transfer. As a result, ${{\rm{MSE}}\left( {\tilde {\cal I}_{TAB}^{hc},\tilde {\cal I}_{SAB}^{hc}} \right)}$ and ${{\rm{LPIPS}}\left( {\tilde {\cal I}_{TAB}^{hc},\tilde {\cal I}_{SAB}^{hc}} \right)}$ increase by 30.96\% and 36.56\% in case II. It is worth noting that case II performs well when warping ${\tilde {\cal I}_{SAB}^{hc}}$ to ${\tilde {\cal I}_{SB}^{hc}}$ because warping from ${\tilde {\cal I}_B^{hc}}$ to ${\tilde {\cal I}_{SAB}^{hc}}$ involves less change. This further indicates that case II almost does not change the internal content of the garment. Case III faces an intermediate result supervision problem and cannot prevent the model from shortcutting. Consequently, during training, case III is likely to collapse, resulting in reductions of 14.91\% and 13.84\% in ${{\rm{PSNR}}\left( {\tilde {\cal I}_{SB}^{hc},{\cal I}_B^{hc}} \right)}$ and ${{\rm{IS}}\left( {\tilde {\cal I}_{SB}^{hc},{\cal I}_B^{hc}} \right)}$, respectively.

\begin{figure}[!ht]
\centering
\includegraphics[width=3.2in]{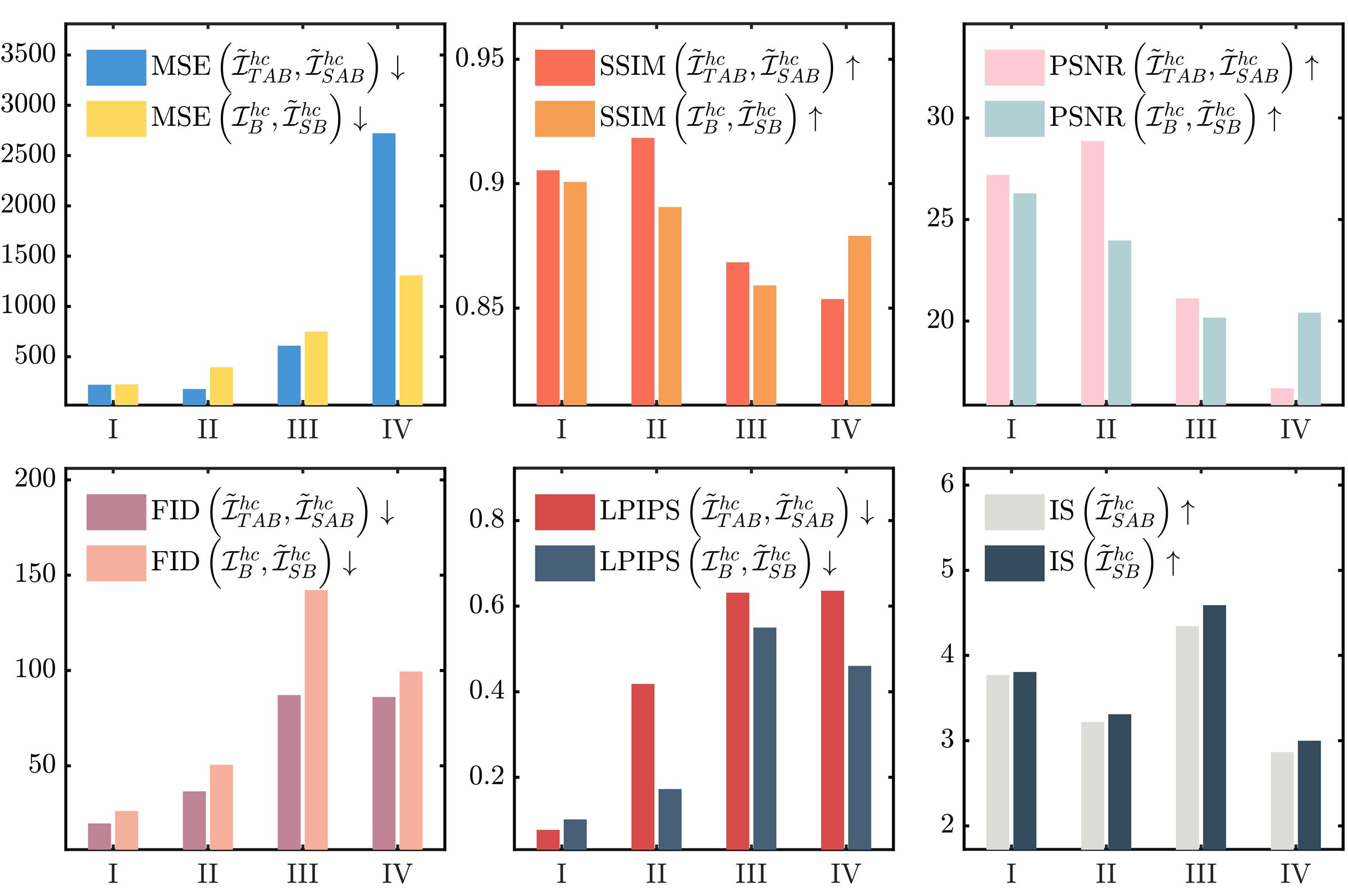}
\caption{Quantitative results in verifying the appropriateness of warping method. Cases I-IV are baseline, mapping, TPS deformation, and TPS deformation + render.}
\label{fig16}
\end{figure}

\begin{figure}[!ht]
\centering
\includegraphics[width=2.6in]{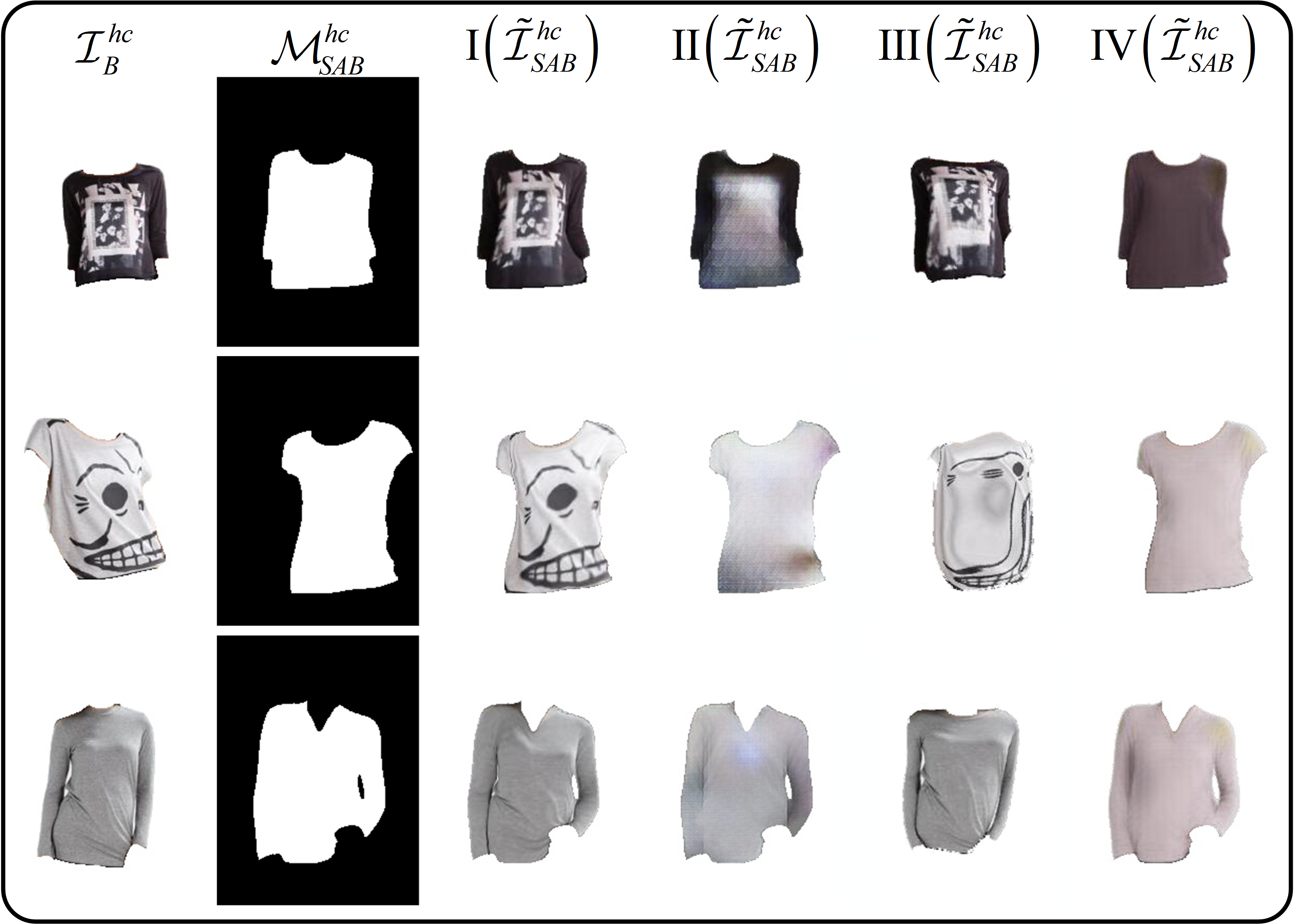}
\caption{Qualitative results in verifying the appropriateness of warping method.}
\label{fig17}
\end{figure}

\textbf{The Appropriateness of Warping Method} As described in {\color{blue} Sec. \ref{sec3.3.1}}, we evaluate the appropriateness of mapping, TPS deformation, and TPS deformation+render for garment transfer. To ensure a fair comparison, we employ the same knowledge distillation pipeline when training these methods. Specifically, for the mapping method, we leverage ${{{\cal T}_w}}$ as the teacher model and the texture module of SwapNet {\color{blue}\citep{raj2018swapnet}} as the student model. The student model is conditioned on ${{\cal M}_{SAB}^{hc}}$ and the garment embedding, which is obtained by ROI pooling of ${{\cal I}_B^{hc}}$; In the case of TPS deformation, we exploit the clothes warping module of ACGPN {\color{blue}\citep{yang2020towards}} as the teacher and student models. The student model is conditioned on ${{\cal M}_B^{hc}}$ and ${{\cal M}_{SAB}^{hc}}$. In the case of TPS deformation + render, we further input the warped result of case III into a refinement network, which is the same as ACGPN. The quantitative and qualitative results are shown in {\color{blue} Fig. \ref{fig16}} and {\color{blue}\ref{fig17}}.

Case II is burdened with complex texture mapping due to variations in the texture and decoration of the garment. Consequently, an average color is padded into the target shape, resulting in blurred texture and discarded decoration. This causes an increase of 85.54\% and 92.51\% in ${{\rm{FID}}\left( {\tilde {\cal I}_{TAB}^{hc},\tilde {\cal I}_{SAB}^{hc}} \right)}$ and ${{\rm{FID}}\left( {\tilde {\cal I}_{SB}^{hc},{\cal I}_B^{hc}} \right)}$, respectively. As TPS deformation has a low DOF, TPS deformation is skilled in warping a coarse result for virtual try-on garments with standard poses and large scales, but it is hard to tackle the fine warping from pose to pose for garment transfer. Therefore, case III cannot accurately warp the garment to match the target shape and tends to produce distortion, leading to an increase of 175.44\% and 341.51\% in ${{\rm{MSE}}\left( {\tilde {\cal I}_{TAB}^{hc},\tilde {\cal I}_{SAB}^{hc}} \right)}$ and ${{\rm{FID}}\left( {\tilde {\cal I}_{TAB}^{hc},\tilde {\cal I}_{SAB}^{hc}} \right)}$, respectively. In case IV, rendering texture is also a challenge, and the poor warped result and unaligned ${{\cal I}_B^{hc}}$ cause training instability and collapse, resulting in a reduction of ${{\rm{IS}}\left( {\tilde {\cal I}_{SAB}^{hc}} \right)}$ by 24.01\%.

\begin{figure}[!ht]
\centering
\includegraphics[width=3.2in]{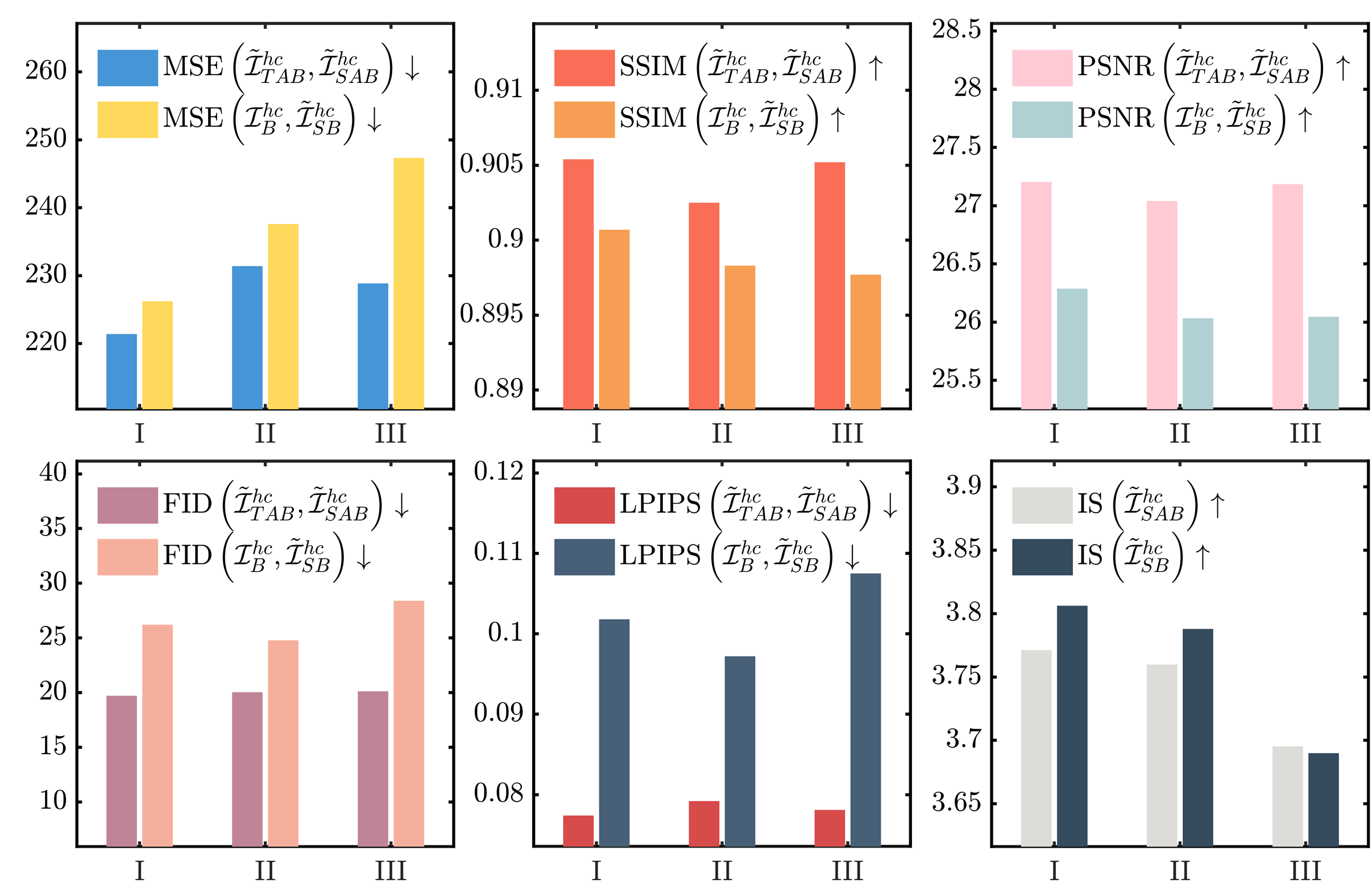}
\caption{Quantitative results in verifying the rationality of phase setting. Cases I-III are baseline, without the initial aligning phase, and with the same architecture as ${{{\cal T}_w}}$}
\label{fig18}
\end{figure}

\begin{figure}[!ht]
\centering
\includegraphics[width=3.3in]{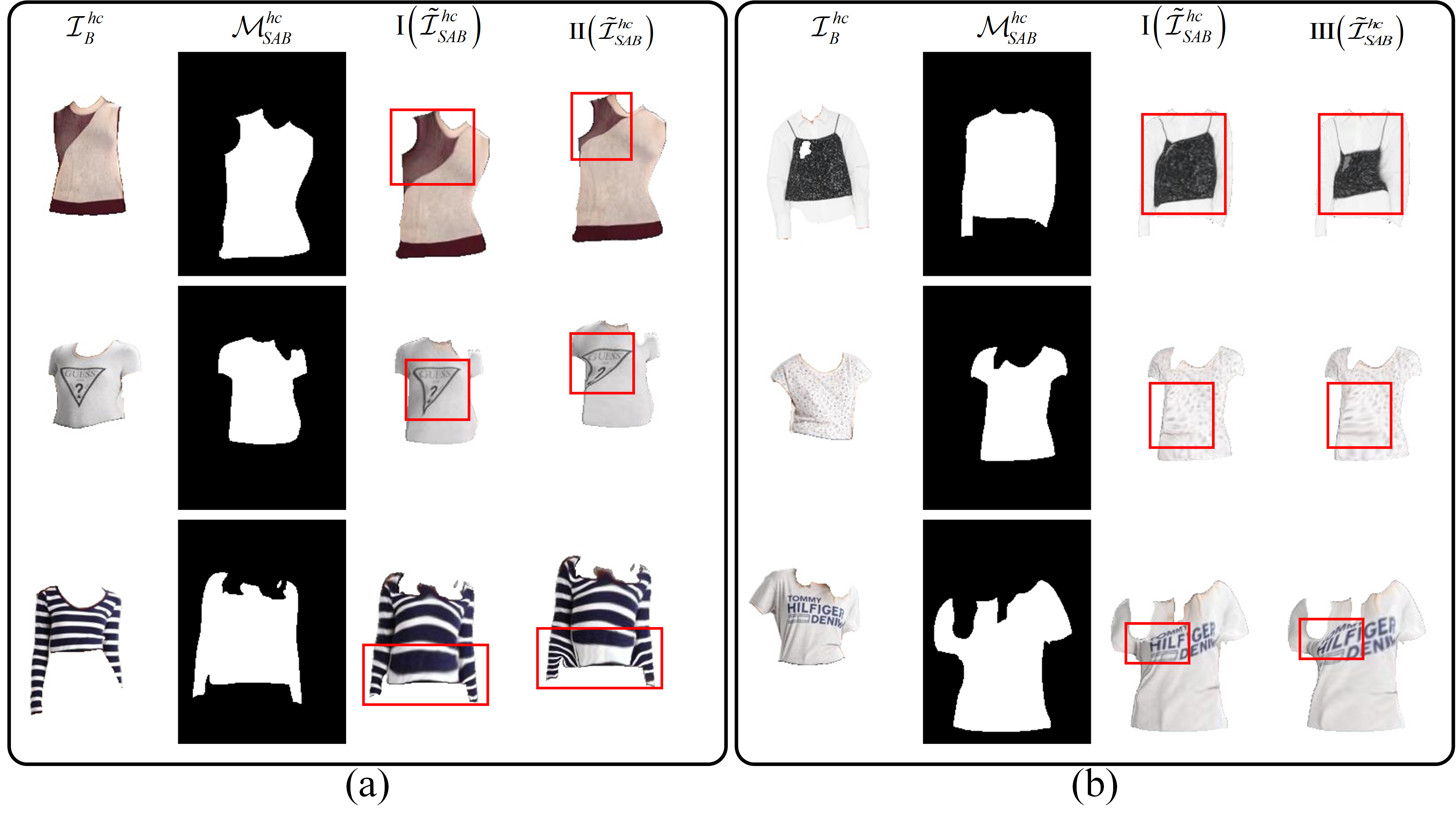}
\caption{Qualitative results in verifying the rationality of phase setting. }
\label{fig19}
\end{figure}

\textbf{The Rationality of Phase Setting} To confirm the effectiveness of the initial aligning phase, we input the unaligned shape pair directly into the flow warping phase. Similarly, to evaluate the architecture of the flow warping phase, we replace it with the same architecture as ${{{\cal T}_w}}$. Quantitative and qualitative results are shown in {\color{blue} Fig. \ref{fig18}} and  {\color{blue} \ref{fig19}}.

In case II, position mapping and shape adjustment are all conducted by high-DOF flow warping.  However, when ${{\cal M}_B^{hc}}$ is significantly smaller than ${{\cal M}_{SAB}^{hc}}$, it tends to excessively stretch the local region to match the target shape precisely. In contrast, case I aligns ${{\cal I}_B^{hc}}$ globally with ${{\cal M}_{SAB}^{hc}}$ to achieve a similar scale and position. Subsequently, flow warping is leveraged to adjust the shape accurately. As a result, the texture blur and garment style variation are reduced, and ${{\rm{MSE}}\left( {\tilde {\cal I}_{TAB}^{hc},\tilde {\cal I}_{SAB}^{hc}} \right)}$ and ${{\rm{LPIPS}}\left( {\tilde {\cal I}_{TAB}^{hc},\tilde {\cal I}_{SAB}^{hc}} \right)}$ decrease by 4.32\% and 2.27\%. In case III, the FSV modulates the pose and garment features to estimate a coarse try-on result, which performs well in virtual try-on without parsing reasoning. However, our garment transfer pipeline provides the fined target shape, rendering the mechanism of FSV unnecessary. Conversely, the innovation of case I learns correspondence at the shape and content level, emphasizing semantic alignment and consistency in garment warping. As a result, ${\tilde {\cal I}_{SAB}^{hc}}$ is more realistic, and ${{\rm{IS}}\left( {\tilde {\cal I}_{SAB}^{hc}} \right)}$ and ${{\rm{IS}}\left( {\tilde {\cal I}_{SB}^{hc}} \right)}$ increase by 2.05\% and 3.15\%.

\subsubsection{Arm Regrowth}
\label{sec4.2.3}
\begin{figure}[!ht]
\centering
\includegraphics[width=3.3in]{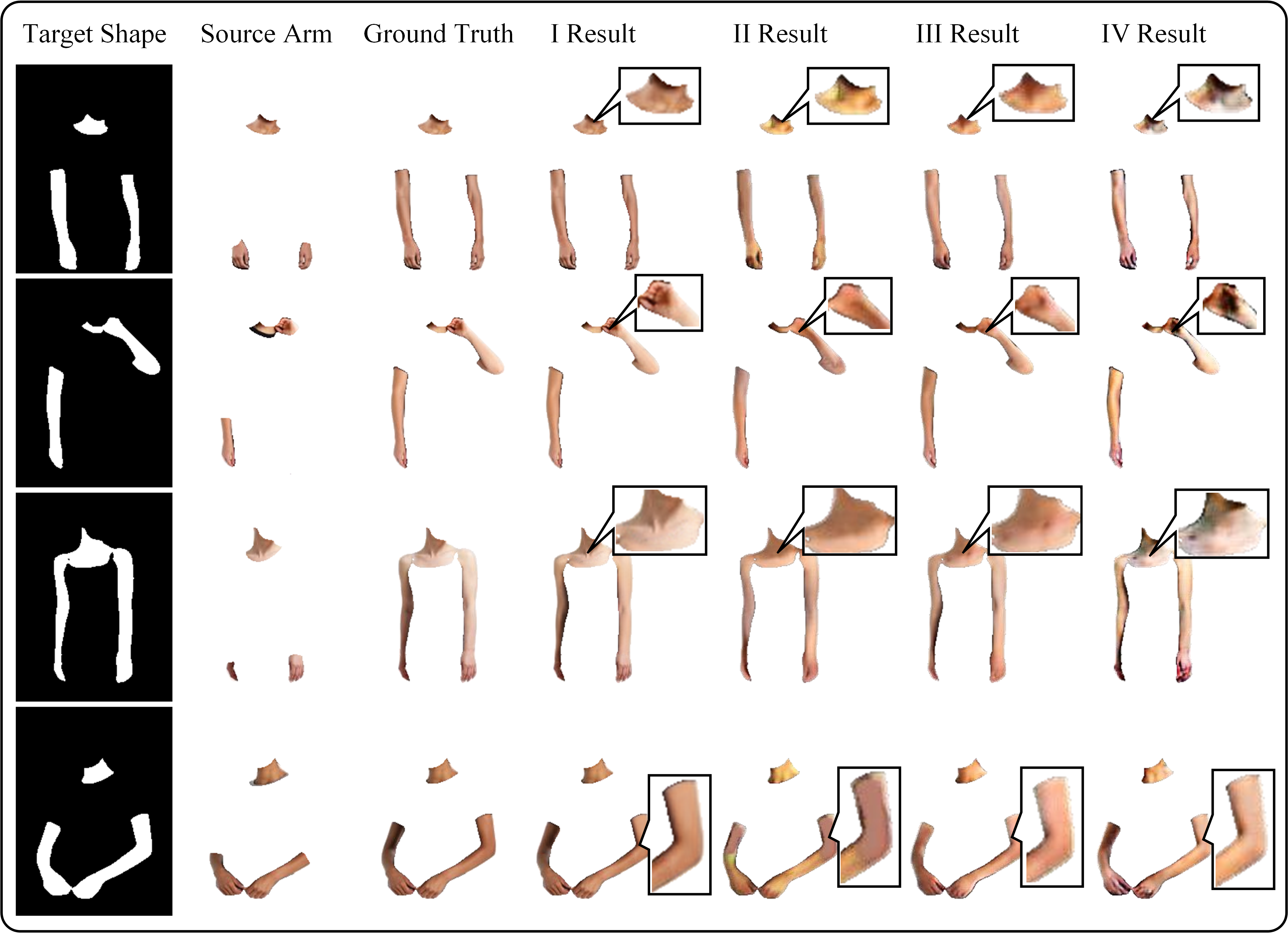}
\caption{Qualitative results in arm regrowth. Cases I-IV are baseline; affine transformation; affine transformation and cropping, affine transformation, cropping and flipping.}
\label{fig20}
\end{figure}

In this section, we aim to demonstrate the effectiveness of our training strategy for arm regrowth. As described in {\color{blue} Sec. \ref{sec3.4}}, due to the unavailability of ground truth data for arm regrowth, we can only train ${{G_r}}$ in a self-supervised manner when ablating our training strategy. Specifically, similar to {\color{blue} Sec. \ref{sec4.2.1}}, we employ three operation combinations to augment ${{\cal I}_A^{ha}}$ during the training phase, and ${{G_r}}$ is conditioned on the augmented ${{\cal I}_A^{ha}}$ and ${{\cal M}_A^{ha}}$ to predict ${\tilde {\cal I}_A^{ha}}$. During the testing phase, all models are conditioned on ${{\cal I}_A^{ha}}$, ${{\cal M}_{SAB}^{ha}}$, and ${{\cal M}_A^{ha}}$ and follow the data processing method in {\color{blue} Sec. \ref{sec3.4}} for the quantitative comparison. The qualitative and quantitative comparisons are shown in {\color{blue} Fig. \ref{fig20}} and {\color{blue} Table \ref{table3}}.

\begin{table}[!ht]
\caption{Quantitative results in arm regrowth.}
\label{table3}
\centering
\begin{tabular}{ccccccc}
\toprule &        MSE	 &PSNR   &SSIM   &FID    &LPIPS  &IS \\ \midrule
          Case I  & 19.63&  36.65&  0.980&   5.96&  0.017& 2.47\\
          Case II &	91.55&	29.28&	0.942&	22.72&  0.055& 2.62\\
          Case III&	74.29&	30.11&	0.949&	16.50&	0.047& 2.55 \\
          Case IV &	88.84&	29.53&	0.949&	20.82&	0.051& 2.62 \\ \bottomrule
\end{tabular}
\end{table}

The task of arm regrowth can be divided into two sub-tasks: propagating the original arm part and reasoning the new arm part. In case II, the augmented ${{\cal I}_A^{ha}}$ is misaligned with ${{\cal M}_A^{ha}}$ through an affine transformation, resulting in ${{G_r}}$ padding the overall target shape with the semantic features of ${{\cal I}_A^{ha}}$. Consequently, the results of case II discard high-frequency details of the hand and neck via inferring rather than content propagating at the overlapped region between ${{\cal M}_{SAB}^{ha}}$ and ${{\cal M}_A^{ha}}$. In case III, new arm exposure is more realistically simulated in the training phase by using random cropping, resulting in a relative reduction of 18.85\% and 14.54\% in MSE and LPIPS respectively when compared to case II. Nevertheless, the results of case III still exhibit artifacts and blurring. Case IV further impedes the propagation of the original arm part by using random flipping, resulting in worse results compared to case III. Unlike these data augmentation techniques, case I aims to enable the input conditions of the training and testing phase to be identical and performs well in both sub-tasks, leading to significant quantitative results.

\subsection{Comparison}
\label{sec4.3}

\subsubsection{Virtual Try-on Method }
\label{sec4.3.1}
\textbf{Baseline} We compare our method with virtual try-on methods in garment transfer. Virtual try-on methods consist of VITON {\color{blue}\citep{han2018viton}}, CP-VTON {\color{blue}\citep{wang2018toward}}, ACGPN {\color{blue}\citep{yang2020towards}}, DCTON {\color{blue}\citep{ge2021disentangled}}, PF-AFN {\color{blue}\citep{ge2021parser}}, and FSV {\color{blue}\citep{he2022style}}.

\begin{figure}[!ht]
\centering
\includegraphics[width=4.5in]{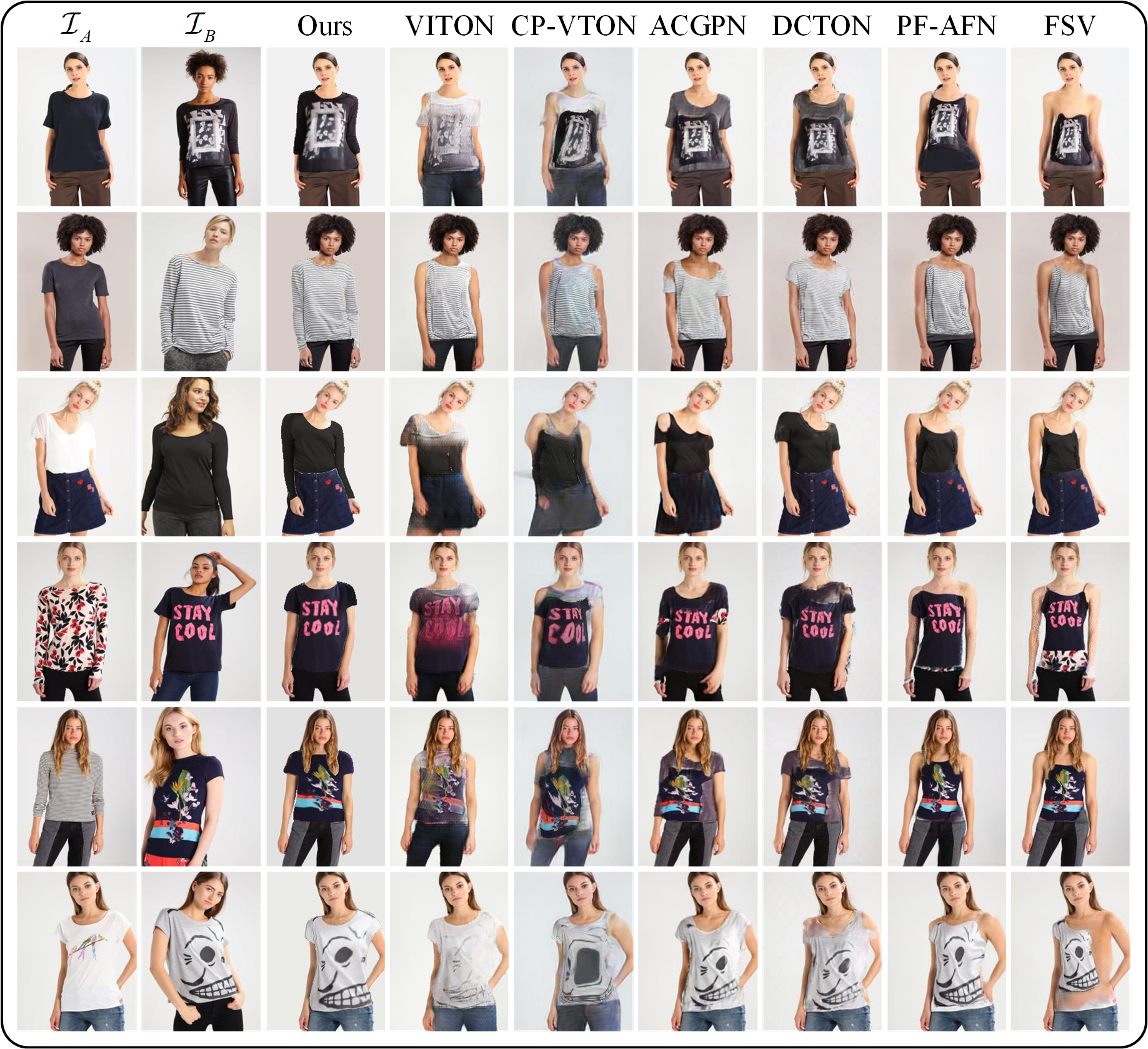}
\caption{Qualitative comparison with virtual try-on methods.}
\label{fig21}
\end{figure}

\textbf{Metrics}  We compare these methods to transfer the garment from ${{{\cal I}_B}}$ to ${{{\cal I}_A}}$ to obtain ${{\tilde {\cal I}_{AB}}}$. In order to perform this transfer, virtual try-on methods are conditioned on both ${{{\cal I}_A}}$ and ${{\cal I}_B^{hc}}$. As the ground truth for ${{\tilde {\cal I}_{AB}}}$ is unavailable, we present several examples in {\color{blue} Fig. \ref{fig21}} and evaluate their subjective image quality using IS {\color{blue}\citep{salimans2016improved}} and hyperIQA {\color{blue}\citep{su2020blindly}} in {\color{blue} Table \ref{table4}}. To conduct a more detailed quantitative comparison, we re-transfer the garment from ${{\tilde {\cal I}_{AB}}}$ to ${{{\cal I}_B}}$ and utilize ${{{\cal I}_B}}$ as the ground truth. We also adopt the same metrics as in {\color{blue} Sec. \ref{sec4.2.2}}. Specifically, we utilize the pre-trained Grapy-ML to extract ${{\tilde {\cal M}_{AB}}}$ from ${{\tilde {\cal I}_{AB}}}$ and yield the transferred garment ${\tilde {\cal I}_{AB}^{hc}}$. Subsequently, virtual try-on methods are conditioned on both ${\tilde {\cal I}_{AB}^{hc}}$ and ${{{\cal I}_B}}$ to predict ${{\tilde {\cal I}_B}}$, and the quantitative results are presented in {\color{blue} Table \ref{table4}}.

\begin{table}[!ht]
\caption{Quantitative comparison with virtual try-on methods. \textbf{Bold} and \underline{underline} highlights represent the best and second-best performance. ${\dag }$ means the unofficial implementation. All these methods adopt official implementations and checkpoints pre-trained on the Zalando dataset except for CP-VTON. As CP-VTON does have an official implementation, we adopt the most starred unofficial code on GitHub ({https:/github.com/sergeywong/cp-vton}).}
\label{table4}
\centering
\resizebox{\linewidth}{!}{
\begin{tabular}{ccccccccc}
\toprule
\multirow{2}{*}{Method} & \multicolumn{2}{c|}{${{{\cal I}_B} \to {{\cal   I}_A}}$} & \multicolumn{6}{c}{${{\tilde {\cal I}_{AB}}   \to {{\cal I}_B}}$}                                     \\ \cmidrule(l){2-9}
                   & IS↑             & \multicolumn{1}{c|}{hyperIQA↑}        & MSE↓           & PSNR↑           & SSIM↑              & FID↓              & LPIPS↓           & IS↑   \\ \midrule
Ours              & \textbf{2.75}     & \textbf{35.94}& \textbf{389.029} & \textbf{23.57}  & \textbf{0.863}     & \textbf{19.14}    & \textbf{0.108}   & \textbf{2.734} \\
VITON             & 2.39             & \underline{31.81}& 3867.18        & 16.15           & 0.700              & 75.91             & 0.258            & 2.374          \\
CP-VTON${\dag }$  & 2.42             & 28.20            & 1853.72        & 16.27           & 0.712              & 53.10             & 0.244            & 2.266          \\
ACGPN             & \underline{ 2.50} & 27.11      & \underline{647.72}  & \underline{21.31}& \underline{0.825} & \underline{27.64} & \underline{0.147}& \underline{2.496}    \\
DCTON             & 2.42              & 29.49           & 770.71         & 20.27          & 0.814               & 32.18             & 0.159             & 2.455          \\
PF-AFN            & 2.31              & 28.80         & 1359.94          & 18.15          & 0.823               & 46.15             & 0.176             & 2.280          \\
FSV               & 2.38              & 27.92          & 1711.46         & 17.12          & 0.820               & 56.50             & 0.191             & 2.317          \\ \bottomrule
\end{tabular}
}
\end{table}

Virtual try-on methods are incapable of capturing the position and style features of transferred garments, resulting in misalignment between warped garments and transfer parsing while yielding artifacts in the unaligned region. In contrast, our method precisely warps ${{\cal I}_B^{hc}}$ towards ${{\cal M}_{SAB}^{hc}}$ through initial aligning and flow warping. Additionally, these virtual try-on methods cannot avoid being affected by the garment shape features of ${{{\cal I}_A}}$ when reasoning a proper transfer parsing, leading to ${{\cal I}_B^{hc}}$ warping toward ${{\cal I}_A^{hc}}$. In contrast, our method infers a parsing that entangles the shape feature of the transfer garment with the pose feature of another person, while maintaining their inherent features. This allows our results to be more aligned with the intended garment transfer. In quantitative comparisons, our method achieves the highest scores of IS and hyperIQA in garment transfer from ${{{\cal I}_B}}$ to ${{{\cal I}_A}}$. Moreover, our MSE and FID are approximately 20.89\% and 34.37\% lower than those of other virtual try-on methods in garment transfer from ${{\tilde {\cal I}_{AB}}}$ to ${{{\cal I}_B}}$.

\subsubsection{Garment Transfer Method}
\label{sec4.3.2}
\textbf{Baseline} We compare our method with other state-of-the-art garment transfer methods, which consist of SwapNet {\color{blue}\citep{raj2018swapnet}}, CT-Net {\color{blue}\citep{yang2021ct}}, and PASTA-GAN {\color{blue}\citep{xie2021towards}}. CT-Net requires paired images of the same person wearing the same garment in different poses for training. Therefore, we utilized its official implementation and pre-trained checkpoint on Deepfashion2 {\color{blue}\citep{ge2019deepfashion2}} and test it on the Zalando dataset.

\begin{figure}[!ht]
\centering
\includegraphics[width=5.4in]{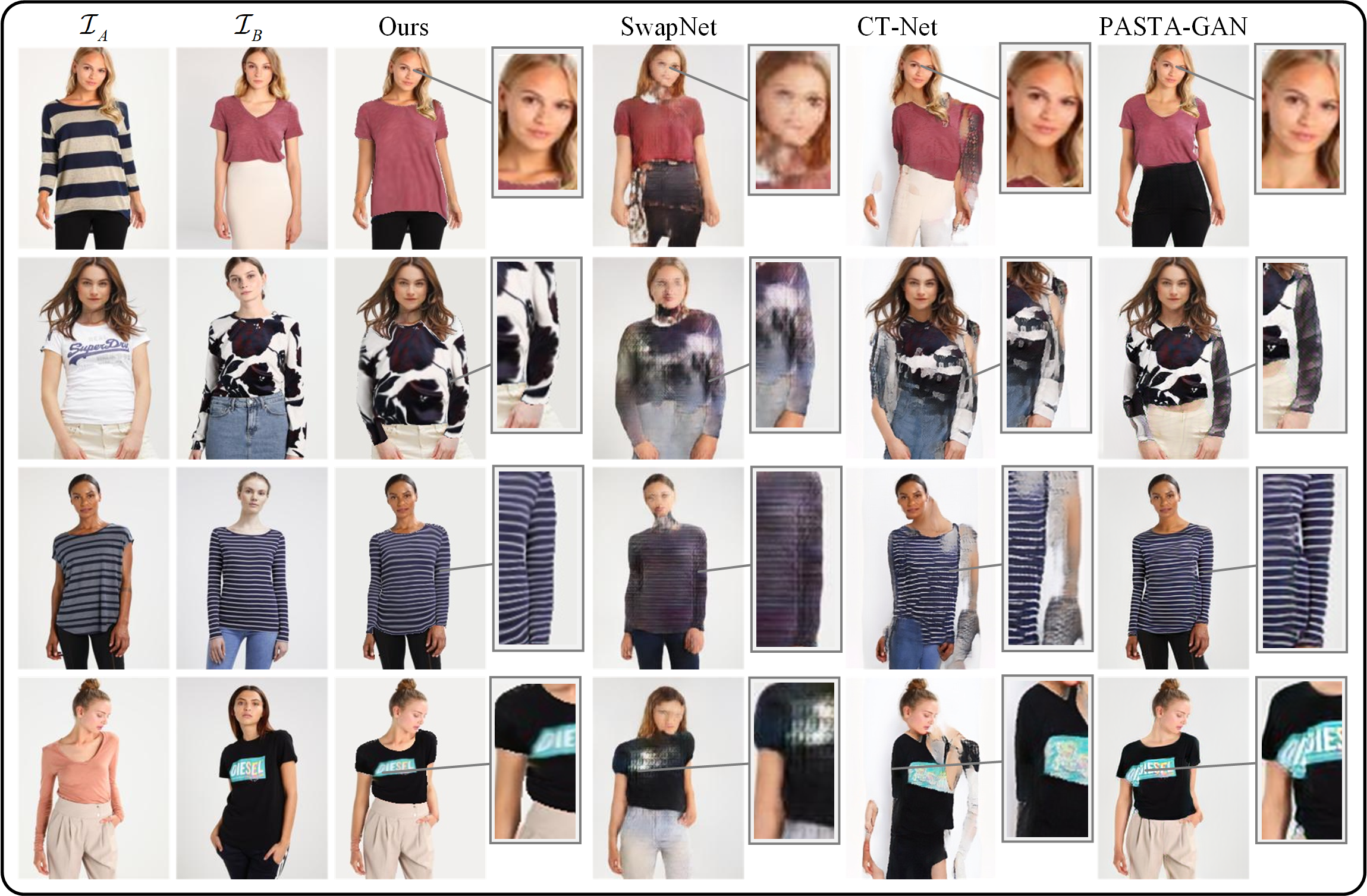}
\caption{Qualitative comparison with garment transfer methods.}
\label{fig22}
\end{figure}

\textbf{Metrics} Similar to {\color{blue} Sec. \ref{sec4.3.1}}, in the garment transfer from ${{{\cal I}_B}}$ to ${{{\cal I}_A}}$, we enumerate some examples and their subjective image quality assessments in {\color{blue} Fig. \ref{fig22}} and {\color{blue} Table \ref{table5}}, respectively. In the garment transfer from ${{\tilde {\cal I}_{AB}}}$ to ${{{\cal I}_B}}$, we provide quantitative results in {\color{blue} Table \ref{table5}}.

\begin{table}[!ht]
\caption{Quantitative comparison with virtual try-on methods. \textbf{Bold} and \underline{underline} highlights represent the best and second-best performance. ${\dag }$ means the unofficial implementation. All these methods adopt official implementations and checkpoints except for SwapNet. As SwapNet does have an official implementation, we adopt the most starred unofficial code on GitHub ({https://github.com/andrewjong/SwapNet}) and trained it on the Zalando dataset.}
\label{table5}
\centering
\resizebox{\linewidth}{!}{
\begin{tabular}{ccccccccc}
\toprule
\multirow{2}{*}{Method} & \multicolumn{2}{c|}{${{{\cal I}_B} \to {{\cal   I}_A}}$} & \multicolumn{6}{c}{${{\tilde {\cal I}_{AB}}   \to {{\cal I}_B}}$}                                     \\ \cmidrule(l){2-9}
                        & IS↑      &\multicolumn{1}{c|}{hyperIQA↑}        & MSE↓              & PSNR↑            & SSIM↑            & FID↓             & LPIPS↓           & IS↑            \\ \midrule
Ours             & \textbf{2.75}   & \textbf{35.94}  & \textbf{389.029}  & \textbf{23.57}   & \textbf{0.863}   & \underline{19.14}& \underline{0.108}& \textbf{2.734} \\
SwapNet${\dag }$ & \underline{2.73}& 27.20           & 1705.66           & 16.31            & 0.677            & 67.45            & 0.269            & 2.557          \\
CT-Net           & 2.48            &\underline{33.24}& 2980.83           & 13.86            & 0.626            & 86.69            & 0.352            & 2.266          \\
PASTA-GAN        & 2.71            & 32.93           & \underline{621.76}& \underline{21.07}& \underline{0.845}& \textbf{15.03}   & \textbf{0.105}   &\underline{2.725}\\ \bottomrule
\end{tabular}
}
\end{table}

SwapNet learns statistical embedding for each category and paints the target shape at a feature level, imposing a heavy learning burden on the CNN model. As shown in the first row of {\color{blue} Fig. \ref{fig22}}, when handling in local region with complex textures or structures, like faces, SwapNet is prone to miss high-frequency details and blur texture, where hyperIQA is decreased by 24.31\%. CT-Net treats the garment transfer as the pose transfer task and implements it in a self-supervision manner. As shown in the second row of {\color{blue} Fig. \ref{fig22}}, in case of no retraining on the Zalando dataset, CT-Net faces the generation and adaption problems in the testing phase, where the pose feature is misaligned and arm is not exposed in transferring a garment with a different type. PASTA-GAN disentangles pose features from garments via patch division and trains a garment transfer model to re-entangle them in a self-supervision manner. As shown in the third and fourth rows in {\color{blue} Fig. \ref{fig22}}, patch division leaves apparent boundaries and makes the re-entangled garment discontinuous across patches, decreasing SSIM by 10.61\%. In contrast, our method leverages a warping mechanism to change the pose and handle arm exposure via an explicit regrowth mechanism, achieving the highest scores of IS and hyperIQA among the garment transfer methods discussed.

\subsection{Failure Case}
\begin{figure}[!ht]
\centering
\includegraphics[width=2.8in]{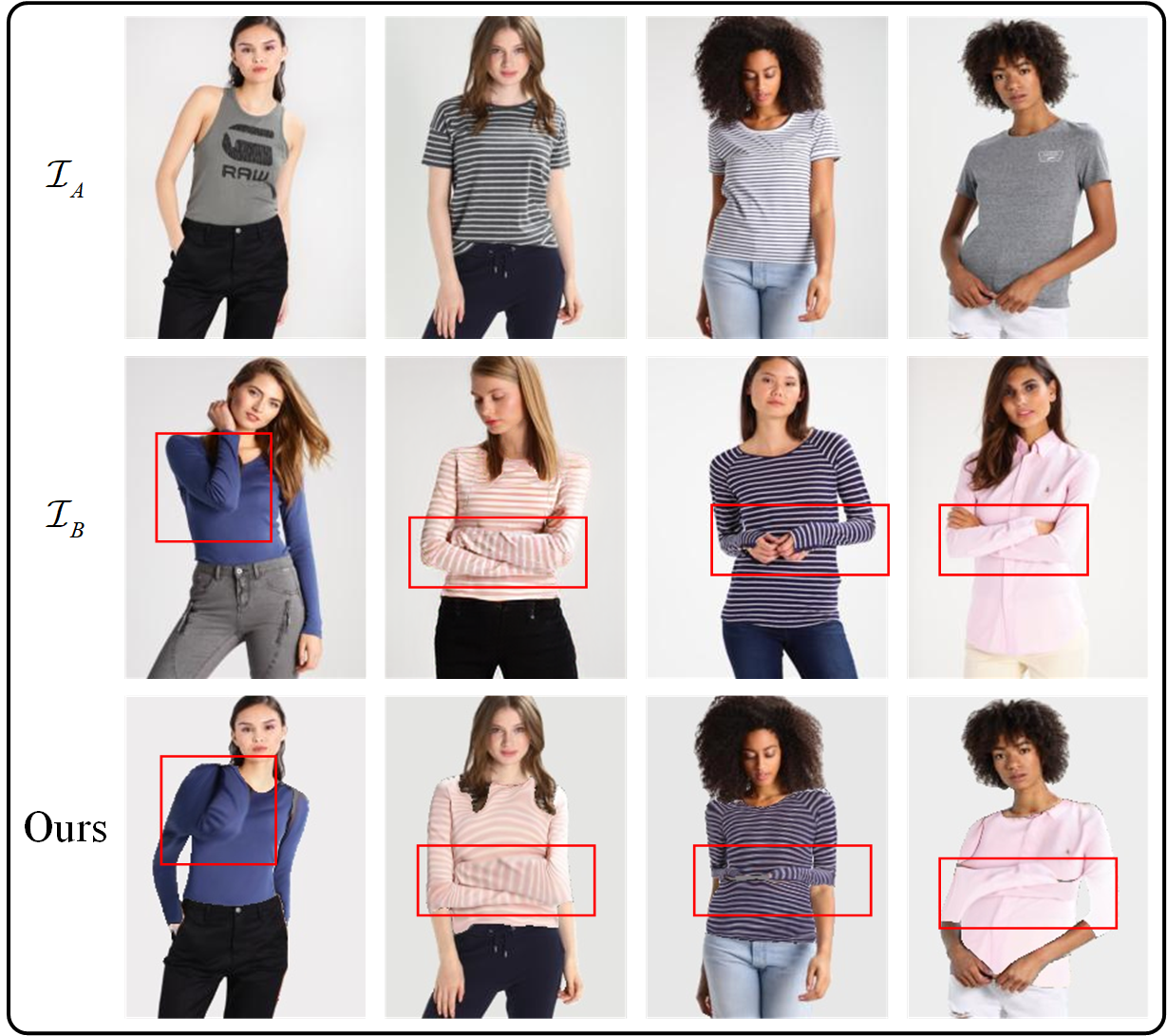}
\caption{Failure cases of our method.}
\label{fig23}
\end{figure}

Our method has limited performance in transferring a garment with the self-occlusion phenomenon. As {\color{blue} Fig. \ref{fig23}} shows, this phenomenon easily appears when a person crosses or raises her arms. With the continuity property in our garment warping, it is hard to tear down the self-occlusion region and leave a blank, thus, the sleeve is prone to be put in an incorrect position. To avoid these failure cases, a straightforward trick is to not take these self-occlusion images as inputs, and for future technical progress, we will focus on establishing the explicit mapping relationship of garments with different poses, via the 3D dense pose or human body model, and explore to employ inpainting mechanism to infer the unobserved region.

\subsection{Generalization}
\begin{figure}[!ht]
\centering
\includegraphics[width=5.4in]{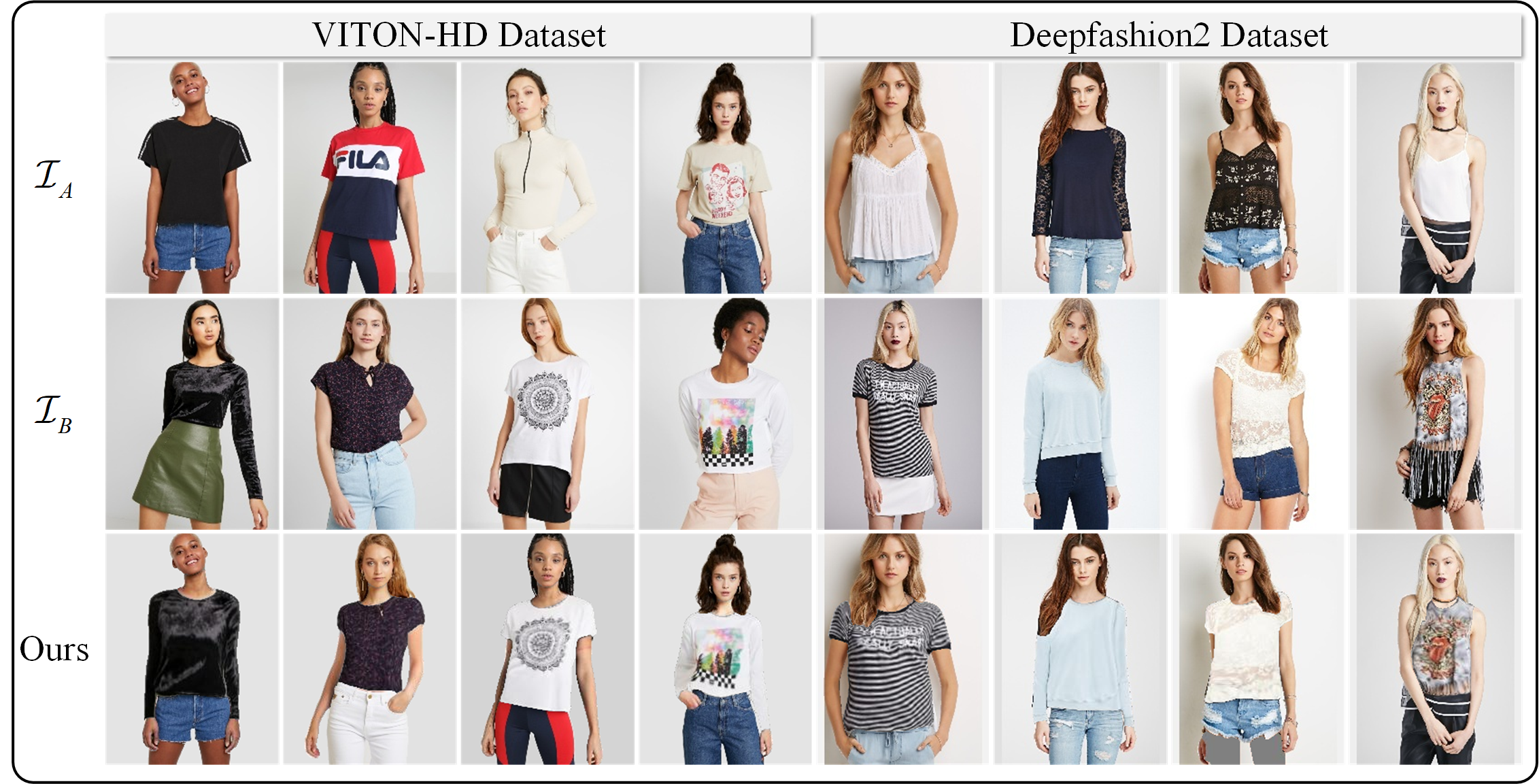}
\caption{Qualitative examples of our method on VITON-HD and Deepfashion2 datasets.}
\label{fig24}
\end{figure}

To verify the generalization capacity, we evaluate our method on VITON-HD {\color{blue}\citep{choi2021viton}} and deepfashion2 {\color{blue}\citep{ge2019deepfashion2}} datasets in case of not retraining our method on them. The qualitative examples and quantitative results are shown in {\color{blue} Fig. \ref{fig24}} and {\color{blue} Table \ref{table6}}. Experimental results show that our method still can robustly generate high-quality transferred images. It demonstrates that transfer parsing reasoning and garment warping mechanisms are available for general cases, and our method can be applied in real transfer scenes.

\begin{table}[!ht]
\caption{Quantitative results of our method VITON-HD and Deepfashion2 datasets.}
\label{table6}
\centering
\begin{tabular}{ccc}
\toprule                & IS↑	  & hyperIQA↑   \\ \midrule
          VITON-HD      & 2.98    & 38.32\\
          Deepfashion2  & 2.56    &	40.06\\ \bottomrule
\end{tabular}
\end{table}

\section{Conclusion}
\label{sec5}
This paper proposes a novel garment transfer method supervised with knowledge distillation from virtual try-on. Our method first reasons the transfer parsing to provide shape prior to downstream tasks. We employ a multi-phase teaching strategy to supervise the training of the transfer parsing reasoning model, learning the response and feature knowledge from the try-on parsing reasoning model. To correct the teaching error, it transfers the garment back to its owner to absorb the hard knowledge in the self-study phase. Guided by the transfer parsing, we adjust the position of the transferred garment via STN to prevent distortion. Afterward, we estimate a progressive flow to precisely warp the garment with shape and content correspondences. To ensure warping rationality, we supervise the training of the garment warping model using target shape and warping knowledge from virtual try-on. To better preserve body features in the transfer result, we propose a well-designed training strategy for the arm regrowth task to infer new exposure skin. Experiments demonstrate that our method has state-of-the-art performance compared with other virtual try-on and garment transfer methods in garment transfer, especially for preserving garment texture and body features.

\section{Declaration of competing interest}
\label{sec6}
The authors declare that they have no known competing financial interests or personal relationships that could have appeared to influence the work reported in this paper.

\section{Acknowledgements}
\label{sec7}
This research was supported by the National Key R\&D Program of China (No. 2018YFB1700700).

\bibliographystyle{elsarticle-num-names}
\bibliography{main}

\end{document}